\pdfoutput=1

\documentclass[11pt]{article}

\PassOptionsToPackage{table,xcdraw}{xcolor}
\usepackage[preprint]{acl}

\usepackage{times}
\usepackage{latexsym}

\usepackage[T1]{fontenc}

\usepackage[utf8]{inputenc}

\usepackage{microtype}

\usepackage{inconsolata}

\usepackage{graphicx}
\usepackage{booktabs}
\usepackage{multirow}
\usepackage{hyperref}

\usepackage{amsmath}
\usepackage[table]{xcolor}
\usepackage{authblk}
\usepackage{listings} 
\usepackage{tikz}
\newcommand*\circled[1]{\tikz[baseline=(char.base)]{
    \node[shape=circle,draw,inner sep=1pt,fill=black,text=white] (char) {#1};}}

\title{GQSA: Group Quantization and Sparsity for Accelerating Large Language Model Inference}

\author{
Chao Zeng\textsuperscript{*}, Songwei Liu\thanks{These authors contributed equally.} \thanks{Project Leader.}, Shu Yang, Fangmin Chen, Lean Fu, Xing Mei \thanks{Corresponding author} \\
ByteDance Inc, \\
\texttt{\{zengchaocs, cfangmin\}@gmail.com, \{liusongwei.zju, xing.mei\}@bytedance.com} \\
}

\begin{document}
\maketitle
\begin{abstract}
Model compression has emerged as a mainstream solution to reduce memory usage and computational overhead. This paper proposes \textbf{GQSA}, a novel model compression framework specifically designed for LLMs. Traditional methods typically focus exclusively on either quantization or sparsification, but relying on a single strategy often results in significant performance loss at high compression rates. In contrast, GQSA integrates quantization and sparsification in a tightly coupled manner, leveraging GPU-friendly structured group sparsity and quantization for efficient acceleration. Building upon system-algorithm co-design principles, we propose a two-stage sparse optimization strategy that ensures the performance superiority of the compressed model. On the engine side, we introduce a "task-centric" parallel strategy, which, to the best of our knowledge, is the first application in the domain of sparse computing. Compared to the traditional 2:4 sparse method, the GQSA offers a more flexible and adjustable sparsity rate, as well as a higher weight compression rate, and is efficiently compatible with weight-only quantization methods. Experimental results demonstrate that, under the GQSA W4S50\% compression setting, the model’s accuracy surpasses that of both 2:4 pruning and W2 quantization. Furthermore, at the inference level, GQSA outperforms W2 by 1.26$\times$ and 2:4 pruning by 2.35$\times$ in terms of speed. 
\end{abstract}

\section{Introduction}
\label{introduction}
Sparsity, combined with quantization~\cite{lin2024awq,shao2023omniquant}, is a powerful approach to enhance model inference performance, reduce the size of LLMs, and enable their deployment on edge devices such as PCs~\cite{gu2024minillm,liu2022multi}. However, current sparsification strategies exhibit limited acceleration benefits due to the unstructured sparsity patterns typically generated by existing unstructured pruning methods~\cite{ han2015learning, sun2023simple}, which are poorly suited for hardware acceleration. Strategies such as SparseGPT~\cite{frantar2023sparsegpt} and Wanda~\cite{sun2023simple} address this issue by adopting a 2:4 sparsity pattern, leveraging NVIDIA GPUs' Sparse Tensor Core units for acceleration. Nevertheless, these approaches are constrained by hardware requirements such as a minimum operation shape of [m, n, k] = [16, 8, 16], which restrict their applicability to compute-intensive tasks like GEMM operations. These limitations pose significant challenges in accelerating the decoding process, the primary performance bottleneck in LLMs~\cite{zeng2024abq}. Unlike GEMM, decoding involves GEMV operations, where the Tensor Core's compute resources are underutilized, with approximately 87.5\% of resources being wasted~\cite{mishra2021accelerating}. Consequently, SparseGPT and Wanda achieve up to 50\% sparsity but remain inefficient in practical scenarios. Furthermore, these methods are incompatible with weight-only quantization, as Sparse Tensor Cores require both weights and activations to are floating-point or integer formats. Combining sparsification with weight-activation quantization leads to excessive compression of activation value representation, resulting in severe performance degradation. This limitation significantly diminishes the practical utility of existing sparsification strategies.

To address these challenges, we propose a novel model compression method called GQSA, designed specifically for the decoding process and efficiently compatible with weight-only per-group quantization. GQSA explores a group sparsity pattern beyond the conventional 2:4 sparsity, achieving a better trade-off between accuracy and speed through a combination of algorithm-level optimizations and a customized software engine. Specifically, we reinterpret weight pruning as a particular form of quantization and introduce a group pruning based on group quantization. Our method incorporates the Block Sparse Row (BSR) format and designs a compact, low-precision weight storage structure to maximize the compression benefits of pruning and quantization. The GQSA method consists of two main stages. The first stage, Block Quantization-Pruning Optimization (BQPO), calibrates model parameters at the block level by optimizing weight distributions within block to minimize performance loss caused by group pruning and quantization. In the second stage, End-to-End Optimized Quantization-Pruning (E2E-OQP), the backbone network’s weights are frozen, and only the quantization parameters are fine-tuned to optimize the global network performance. Unlike BQPO, E2E-OQP considers the global error distribution across blocks. Freezing the backbone network can not only reduce memory usage but also improve training efficiency. Extensive experiments demonstrate that GQSA achieves significant advantages in both model accuracy and inference speed, especially when applied to newly released advanced models such as LLaMA-3 and LLaMA-3.1 model family and Qwen2.5 models.

In summary, our contributions are as follows. 
\begin{itemize}
\item We propose a sparse scheme seamlessly compatible with widely used weight-only and weight-activation quantization, effectively accelerating GEMV operations and reducing memory usage.
\item We introduce a task-centric parallel implementation, addressing the workload balancing issue in sparse acceleration.
\item We integrate group pruning with low-bit quantization techniques and achieves outstanding model performance through the two-stage optimization process of BQPO and E2E-OQP.
\end{itemize}

\section{Related work}
\textbf{Compressing Large Language Models.} Pruning and quantization are the two primary techniques for compressing LLMs. Pruning methods can be classified into structured~\cite{chen2024compressing, ma2023llm,ashkboos2024slicegpt}, semi-structuredcite~\cite{frantar2023sparsegpt, sun2023simple, fang2024maskllm}, and unstructured~\cite{han2015deep_compression, han2015learning, sun2023simple} pruning, depending on the granularity of pruning. Structured pruning operates at a coarser granularity and offers significant acceleration, but it often results in a substantial loss of accuracy~\cite{wang2024moreaupruner}, limiting its application in LLMs. Unstructured pruning better preserves accuracy but provides limited improvements in inference speed~\cite{frantar2023sparsegpt}. Semi-structured pruning strikes a balance between accuracy retention and acceleration, though it is constrained by a sparsity of 50\%, reducing its flexibility. Quantization reduces model size by replacing floating-point numbers with low-precision integers, which accelerates memory access during inference. Currently, high-bit quantization techniques such as AWQ~\cite{lin2024awq}, GPTQ~\cite{frantar2022gptq}, QuIP~\cite{chee2024quip}, OmniQuant~\cite{shao2023omniquant}, and OWQ~\cite{lee2024owq} are widely adopted. However, extremely low-bit quantization poses significant challenges, with mainstream methods struggling to maintain performance at low-bit levels. While techniques like AQLM~\cite{egiazarian2024extreme} and QuIP\#~\cite{tseng2024quip} aim to enhance low-bit quantization, they rely on vector quantization and complex codebooks, which hinder inference acceleration. Overall, existing model compression techniques continue to face substantial challenges in achieving an optimal balance between flexibility and compression rate.
\begin{figure}[t]
\centering
\includegraphics[width=\linewidth]{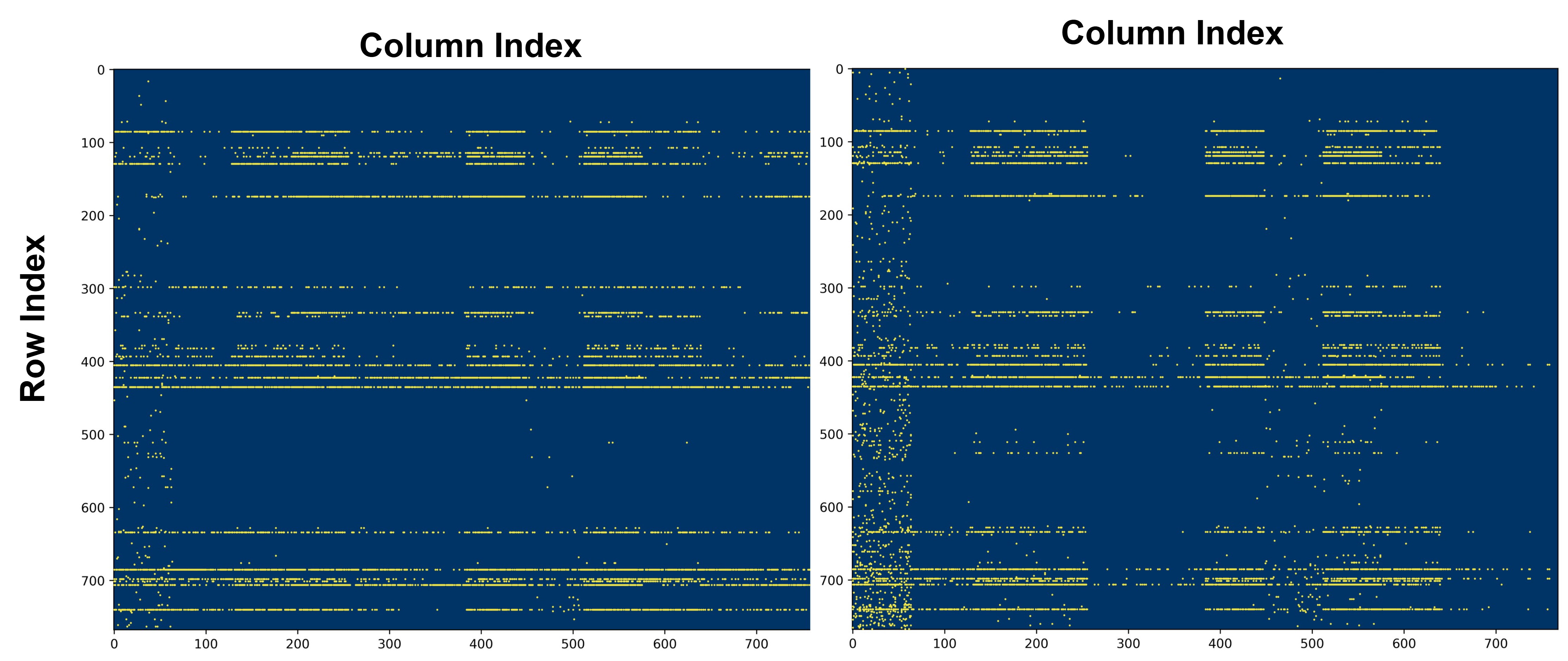}
\vspace{-0.12in}
\caption{The distribution of the top 1\% significant weights in the Hessian matrix, derived from the $k_{proj}$ and $q_{proj}$ distributions in the LLaMA-7B model.}
\vspace{-5pt}
\label{fig:salient_weight}
\end{figure}

\begin{figure*}[ht]
  \centering
  \includegraphics[width=0.85\textwidth]{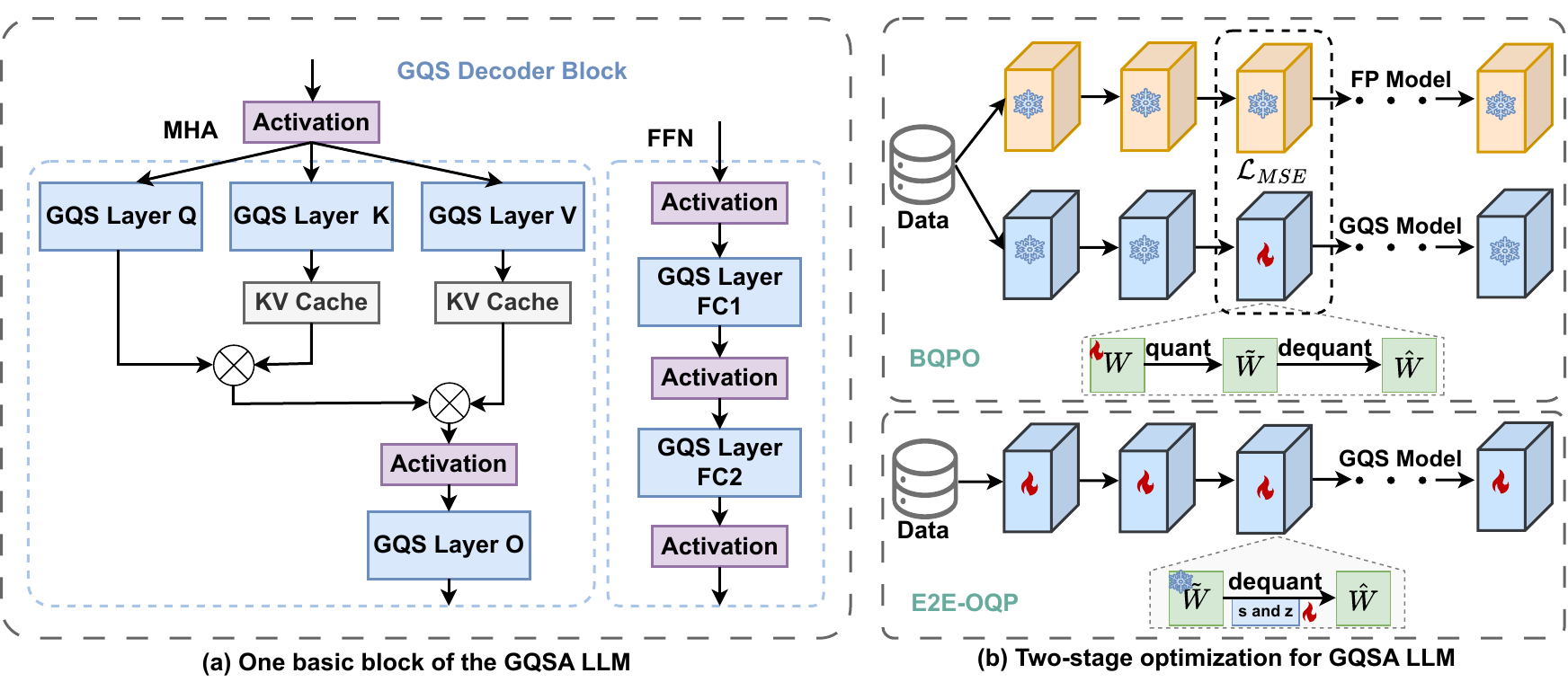}
  \vspace{-0.12in}
  \caption{
  Overview of GQSA. (a) We propose a group quantization and sparse LLMs, where linear layers are replaced by GQS layers. (b) We use the two-stage optimization method BQPO and E2E-OQP to recover the performance of the extremely compressed model.}
  \vspace{-5pt}
  \label{fig:gqs_overview}
\end{figure*}

\noindent \textbf{Advantages of GQSA.} Quantization and sparsity address model redundancy in different ways. Quantization reduces the precision of numerical representations, while sparsity compresses the model by eliminating certain neurons. These two techniques are largely orthogonal, and GQSA leverages both dimensions to achieve flexible and high compression rates. Although both GQSA and 2:4 pruning are semi-structured pruning methods, GQSA offers several advantages over 2:4 pruning. First, GQSA supports an adjustable sparsity rate, whereas 2:4 pruning, designed for NVIDIA's 2:4 TensorCore, mandates a 50\% sparsity rate by forcing two out of every four weights to be zero. Our group sparsity model, in combination with co-designed operators, enables efficient implementation at various sparsity levels. Second, GQSA achieves a higher weight compression rate. For instance, with a 50\% sparsity rate, 2:4 pruning requires additional metadata to identify the positions of retained neurons, which are chosen randomly. In contrast, GQSA stores location information at the group/block level, significantly enhancing compression efficiency. Finally, 2:4 pruning is restricted to NVIDIA's 2:4 TensorCore and is incompatible with mainstream weight-only quantization methods. In contrast, GQSA is highly compatible with weight-only quantization, thanks to its customized two-stage optimization, leading to a substantial increase in overall compression rate.


\section{GQSA}
In this section, we provide a detailed exposition of GQSA. Section~\ref{sec:3.1} begins with an examination of weight quantization and salient weight selection principles. Building upon these foundations, Section~\ref{sec:3.2} introduces the innovative GQS Layer, designed to maximize the compression advantages from both quantization and pruning. The subsequent sections (~\ref{sec:3.3} and ~\ref{sec:3.4}) detail our two-stage optimization algorithm, which delivers exceptional model performance. Concluding the section, ~\ref{sec:3.5} proposes a novel task-centric parallel strategy for efficient inference acceleration.

\subsection{Preliminary} \label{sec:3.1}
\textbf{Weight Quantization.} 
LLM quantization maps floating-point values to a lower-bit discrete value space, significantly reducing model size, enhancing computational efficiency, and accelerating inference. The process typically involves two steps: determining the quantization parameters (scale and zero-point) and computing the corresponding quantized tensor. For uniform asymmetric quantization, which is used in this paper, the scale $s$ and zero-point $z$ are determined by:
\begin{equation}\label{eq:qparameter}
\begin{aligned}
\scalebox{1}{$
s=\frac{\mathrm{max}(W)-\mathrm{min}(W)}{2^{n}-1 } , z=- \left \lfloor \frac{\mathrm{min}(W)}{s}  \right \rceil ,
$}
\end{aligned}
\end{equation}
where $W$ represents the model weights and $n$ denotes the quantization bit-width. The elements of the quantized tensor can be computed as follows:
\begin{equation}\label{eq:quant}
\begin{aligned}
\tilde{W} = \mathrm{clamp}(\left \lfloor \frac{W}{s}  \right \rceil + z, 0, 2^{n}-1),
\end{aligned}
\end{equation}
where $\left \lfloor \cdot  \right \rceil$ represents the rounding operation, and $\tilde{W}$ represents the quantized integer weights. When it is necessary to update the quantized weights of the model, the weights are converted back to full precision during the forward propagation phase, as shown below:
\begin{equation}\label{eq:dequant}
\begin{aligned}
\hat{W} = (\tilde{W}-z )\cdot s ,
\end{aligned}
\end{equation}
where $\hat{W}$ denotes the dequantized weights utilized in the forward computation. The processes of quantization (as shown in Equation~\eqref{eq:quant} and dequantization (as shown in Equation~\eqref{eq:dequant} are integrated into the computational graph, enabling quantization-aware optimization through gradient descent.

\noindent \textbf{Salient Weight.} In LLMs, different weights exhibit different importance. By pruning unimportant weights, memory usage can be greatly reduced while maintaining nearly unchanged performance. Early studies used the absolute value of weights to evaluate weight importance, but ignored the role of activation. The Hessian metric combines weights and activations and is a more effective metric that has been verified by multiple methods~\cite{shang2023pb,frantar2023sparsegpt}. Therefore, this paper uses the Hessian matrix to evaluate weight importance.
\begin{equation}\label{eq:hessian}
\begin{aligned}
s_{i} = \frac{w_{i}^{2} }{\left [ H^{-1}  \right ] _{ii}^{2} },  
\end{aligned}
\end{equation}
where $H$ represents the Hessian matrix of each layer, and $w_{i}$ denotes the weight values. In the subsequent sections, $s_{i}$ refers to the criteria for identifying salient weights. AWQ~\cite{lin2024awq} demonstrates that the top 1\% of salient weights in the model are crucial to performance, so accurately retaining these weights is key to performance. Figure~\ref{fig:salient_weight} visualizes the distribution of salient weights in the OPT model, revealing a segmented pattern along the rows. Consequently, group by rows and selecting salient weight group emerges as a natural optimization strategy.

\begin{figure}[th]
  \centering
  \includegraphics[width=\linewidth]{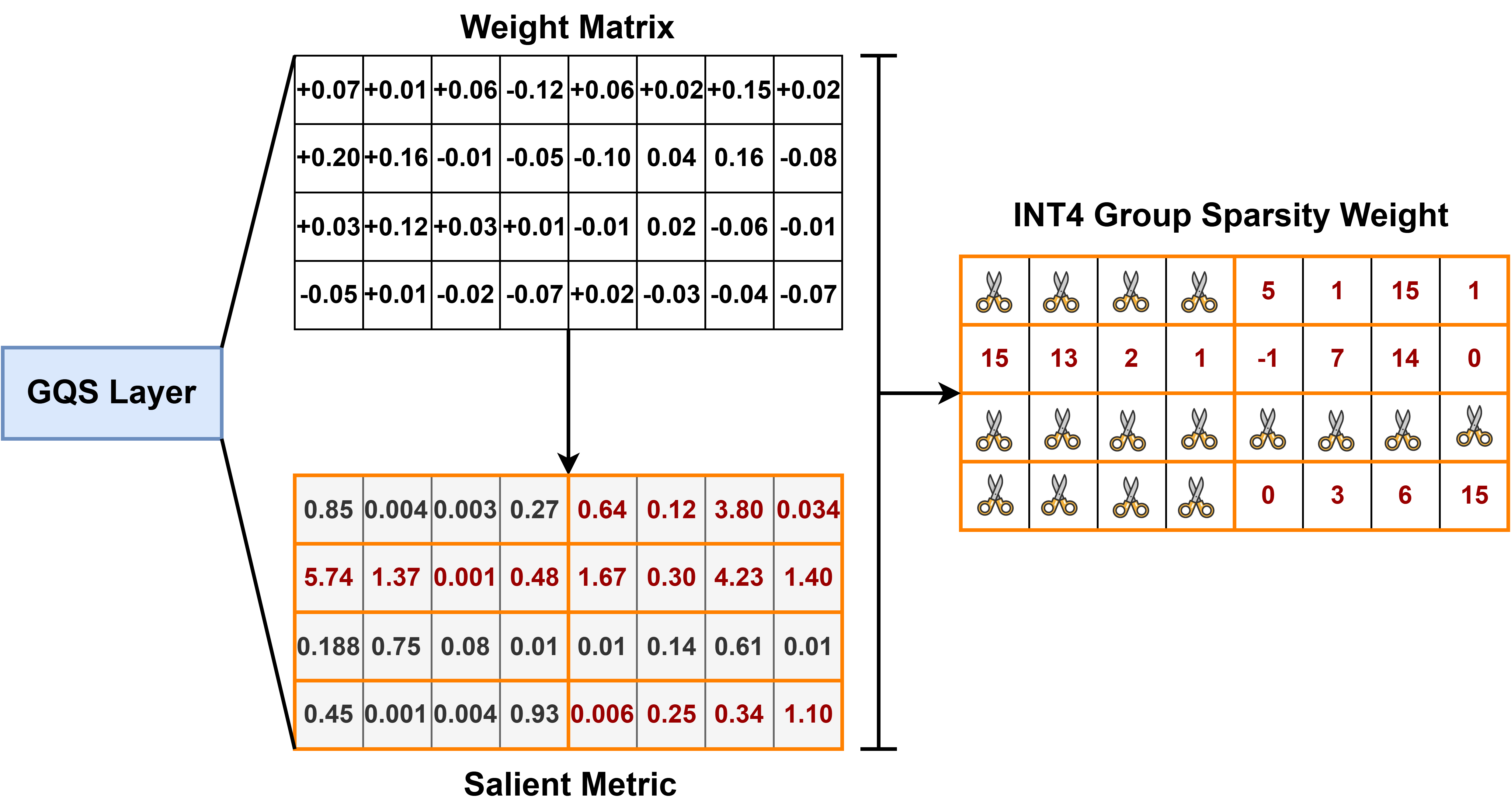}
  \vspace{-7pt}
  \caption{GQSA computes saliency metrics based on weights and activations, grouping the weights along the row dimension (illustrated with groups of four elements). Group pruning is then applied based on the average saliency metrics within each group, resulting in the formation of the GQS layer.}
  \vspace{-7pt}
  \label{fig:gqas}
\end{figure}

\subsection{GQS Layer} \label{sec:3.2}
Weight-only per-group quantization has gained significant recognition in both academia~\cite{lin2024awq,shao2023omniquant,frantar2022gptq} and industry~\cite{gerganov2024llama}. To enable efficient sparse acceleration compatible with weight-only per-group quantization approach, we conduct a comprehensive analysis of the distribution of salient weights within the model. As depicted in Figure~\ref{fig:salient_weight}, we observe that salient weights exhibit distinct segmented distribution patterns. Based on this observation, we introduce a novel structured group pruning method that goes beyond the conventional 2:4 sparsity pattern, leveraging the segmented distribution characteristics of salient weights. As illustrated in Figure~\ref{fig:gqas}, we begin by grouping weights along the row dimension, assuming a group size of 4 for simplicity. For each group, we compute a salient metric using the Hessian matrix. Based on this metric, we prune non-salient weight groups and quantize the remaining salient groups to 4 bits, thereby further compressing the model size. Additionally, by adopting the BSR sparse format, we convert the compression gains from pruning into actual storage savings. The specific storage structure in Figure~\ref{fig:gqas} is shown below:

\begin{lstlisting}
rowIndex = {0, 1, 3, 3, 4}
 groups  = {1, 0, 1, 1}
 values  = {5, 1, 15, 1, 15, 13, 2, 1, 
           -1, 7, 14, 0, 0, 3, 6, 15}
\end{lstlisting}
where \texttt{rowIndex[i]} represent the offset of each row \texttt{i}, where \texttt{i} belongs to the range \([0, \text{rows}]\). The difference \(\texttt{rowIndex[r+1] - rowIndex[i]}\) indicates the number of non-zero groups in the \texttt{i}-th row. Additionally, \(\texttt{rowIndex[rows]}\) represents the total number of non-zero groups. The array \(\texttt{groups[i]}\) stores the indices of the non-zero groups; for instance, if \(\texttt{groups[1] = 0}\), it means that the second group is located in the 0th column (in terms of group units). Finally, \texttt{values} stores the values of the non-zero groups for each row.

\subsection{BQPO} \label{sec:3.3}
In the first stage (Figure~\ref{fig:gqs_overview}(b)), we apply the BQPO method to optimize the GQS model, aiming to mitigate the accuracy degradation caused by group quantization and pruning. This is achieved by adjusting the weight parameters within each block. Traditional QAT methods typically optimize the entire network’s weights in an end-to-end fashion, as illustrated in Equations~\eqref{eq:quant} and~\eqref{eq:dequant}. Similarly, most pruning approaches adopt a global end-to-end strategy to update the remaining unpruned parameters. However, such methods often demand substantial computational resources and large-scale training datasets. To improve optimization efficiency, BQPO adopts a block-wise optimization strategy. Prior studies, such as OmniQuant and AffineQuant, have shown that block-wise optimization can significantly reduce both training time and memory consumption. Unlike OmniQuant and AffineQuant, which primarily optimize quantization parameters (inter-channel smoothing factors and weight clipping thresholds), GQSA suffers from more severe performance degradation due to its combination of high structured sparsity and low-precision quantization. As a result, BQPO focuses on optimizing the remaining weights to recover performance under extreme compression settings. This block-wise approach enables significant performance restoration with minimal additional training cost compared to global optimization techniques.

 \begin{figure}[t]
  \centering
  \includegraphics[width=0.9\linewidth]{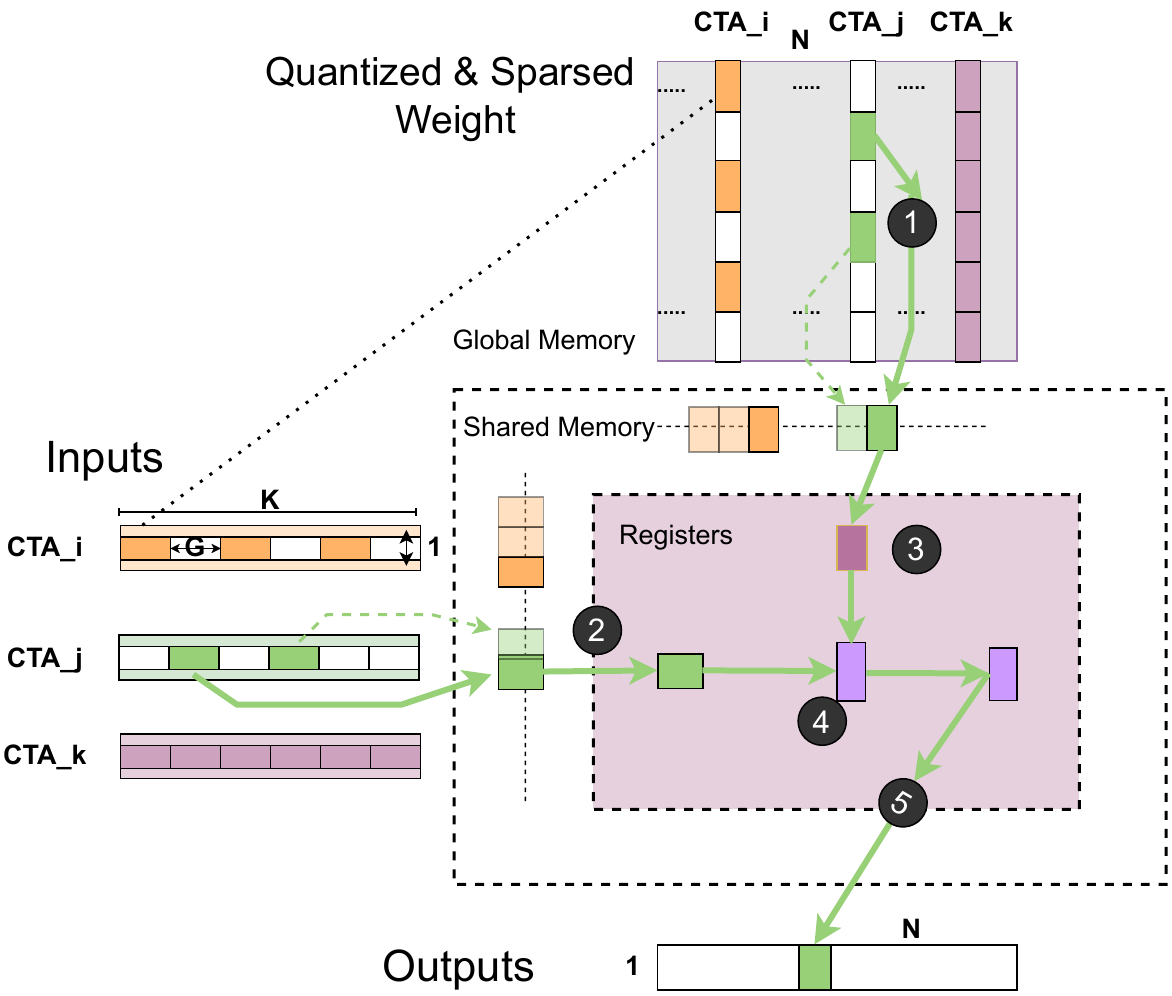}
  \caption{A simplified view of GQSA’s operator calculation flow. G represents sparse and quantized group size.}
  \label{fig:engine}
\end{figure}

\subsection{E2E-OQP} \label{sec:3.4}

Compared to BQPO, E2E-OQP not only performs intra-block optimization but also accounts for the overall error across the entire network, thereby capturing cross-block dependencies. As illustrated in Figure~\ref{fig:gqs_overview}(b), E2E-OQP differs from conventional quantization-aware training (QAT) methods. Assuming that BQPO has already yielded a well-optimized model in the first stage, E2E-OQP initializes training using the BQPO-optimized weights. During this phase, we freeze the primary network weights $\tilde{W}$ and optimize only the quantization parameters $s$ and $z$ to further refine model performance. The design of E2E-OQP underscores the advantages of the GQSA framework. Specifically, during fine-tuning, we employ the block-sparse row (BSR) format: the remaining group weights are quantized to low bit-width and frozen, while pruned groups are discarded entirely. This strategy enables effective fine-tuning of the quantization parameters without requiring sparse masks, thereby restoring the performance of the GQSA model under extreme compression. Overall, E2E-OQP achieves substantial memory savings by focusing solely on the quantization parameters of the remaining groups while maintaining 4-bit quantization across the main network. A detailed comparison of the resource consumption of BQPO and E2E-OQP is provided in Appendix~\ref{ap:train_efficien}, demonstrating the efficiency advantages of the GQSA approach.


\begin{figure}[t]
  \centering
  \includegraphics[width=0.8\linewidth]{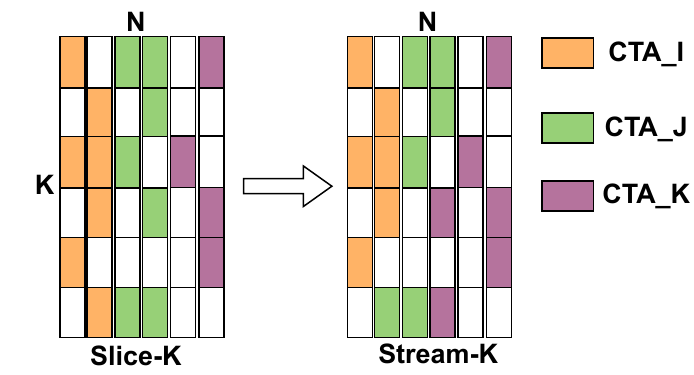}
  \caption{Workload balancing through parallel task partitioning.}
  \label{fig:stream}
\end{figure}

\subsection{Custom Software Engine} \label{sec:3.5}
 GPU has many processing elements called Streaming Multiprocessors (SMs) and uses a large number of threads to perform computing tasks in parallel. Threads are structured into thread blocks (CTAs), which become the smallest scheduling execution unit on SMs. Therefore, the computation target is decomposed and mapped to each thread block, called CTA, to achieve parallel computing. As shown in Figure~\ref{fig:engine}, for a GEMV task of shape 1×N×K, each thread block is responsible for computing a 1×BN output tile, which is decomposed into $\frac{K}{BK}$ sub-GEMV tasks of shape 1×BN×BK. In offline pre-processing, quantized weights are grouped by size G and saved as gguf format along with scaling factors and zero points. This means that each sub-GEMV task computes  $\frac{BK}{G}$ * $BN$ non-sparse groups held by one or more output channels. It should be noted that the logical addresses between non-sparse groups are not necessarily consecutive, so the corresponding activation group needs to be accessed according to the real group index of each group. \circled{1} The thread-block issues asynchronous copy instructions to fetch small chunks of input data (tiles) from global memory to shared memory. \circled{2} As soon as a tile arrives in shared memory, it is further sliced into smaller chunks (fragments) and copied into registers. \circled{3} Once all necessary components are in the registers, the quantized matrix undergoes dequantization. \circled{4} The dequantized matrix and inputs are then processed by TensorCores (MMA) or CudaCores (FMA) instructions. \circled{5} Finally, the accumulated results are written back from the registers to the outputs in global memory. 

\begin{table*}[ht]
\centering
\resizebox{\textwidth}{!}{%
\begin{tabular}{cccccccccccccc}
\toprule
 &  & \multicolumn{2}{c}{LLaMA-7B} & \multicolumn{2}{c}{LLaMA-13B} & \multicolumn{2}{c}{LLaMA-2-7B} & \multicolumn{2}{c}{LLaMA-2-13B} & \multicolumn{2}{c}{LLaMA-3-8B} & \multicolumn{2}{c}{LLaMA-3.1-8B} \\ \cmidrule(l){3-14} 
\multirow{-2}{*}{Setting} & \multirow{-2}{*}{Method} & WikiText2 & C4 & WikiText2 & C4 & WikiText2 & C4 & WikiText2 & C4 & WikiText2 & C4 & WikiText2 & C4 \\ \midrule
 & GPTQ & 44.01 & 27.71 & 15.60 & 15.29 & 36.77 & 33.70 & 28.14 & 20.97 & 210 & 4.1e4 & 250 & 80.3 \\
 & QUIP & 29.74 & 33.74 & 12.48 & 21.94 & 39.73 & 31.94 & 13.48 & 16.16 & 84.97 & 1.3e2 & - & - \\
 & PB-LLM & 24.61 & 49.73 & 17.73 & 26.93 & 25.37 & 29.84 & 49.81 & 19.82 & 44.12 & 79.2 & - & - \\
 & Omniquant & 15.47 & 24.89 & 13.21 & 18.31 & 37.37 & 90.64 & 17.21 & 26.76 & 2.1e4 & 6.0e4 & 7.3e3 & 1.3e4 \\
 & LeanQuant & 15.65 & 17.62 & 9.64 & 10.93 & 16.98 & 17.89 & 10.32 & 11.73 & 41.78 & 36.50 & - & - \\
\multirow{-6}{*}{W2} & SliM-LLM & 14.58 & 32.91 & \textbf{8.87} & 13.85 & 16.01 & 16.00 & 9.41 & \textbf{9.41} & 39.66 & 1.1e2 & - & - \\ \midrule 
2:4 & SparseGPT & \textbf{11.20} & 13.59 & 9.14 & 11.34 & 10.95 & 13.56 & 8.32 & 11.30 & 16.56 & 22.99 & 16.62 & 23.22 \\
2:4 & Wanda & 11.53 & 14.41 & 9.58 & 12.07 & 11.02 & 15.07 & 8.27 & 12.12 & 25.27 & 36.40 & 23.93 & 36.24 \\ \midrule
w4s20\% & \cellcolor[HTML]{EFEFEF} & \cellcolor[HTML]{EFEFEF}6.58 & \cellcolor[HTML]{EFEFEF}8.30 & \cellcolor[HTML]{EFEFEF}5.75 & \cellcolor[HTML]{EFEFEF}7.57 & \cellcolor[HTML]{EFEFEF}6.57 & \cellcolor[HTML]{EFEFEF}8.32 & \cellcolor[HTML]{EFEFEF}5.86 & \cellcolor[HTML]{EFEFEF}7.51 & \cellcolor[HTML]{EFEFEF}8.43 & \cellcolor[HTML]{EFEFEF}12.54 & \cellcolor[HTML]{EFEFEF}8.40 & \cellcolor[HTML]{EFEFEF}12.37 \\
w4s30\% & \cellcolor[HTML]{EFEFEF} & \cellcolor[HTML]{EFEFEF}7.91 & \cellcolor[HTML]{EFEFEF}9.74 & \cellcolor[HTML]{EFEFEF}6.72 & \cellcolor[HTML]{EFEFEF}8.32 & \cellcolor[HTML]{EFEFEF}7.56 & \cellcolor[HTML]{EFEFEF}9.49 & \cellcolor[HTML]{EFEFEF}6.87 & \cellcolor[HTML]{EFEFEF}8.49 & \cellcolor[HTML]{EFEFEF}9.79 & \cellcolor[HTML]{EFEFEF}14.58 & \cellcolor[HTML]{EFEFEF}9.69 & \cellcolor[HTML]{EFEFEF}14.32 \\
w4s40\% & \cellcolor[HTML]{EFEFEF} & \cellcolor[HTML]{EFEFEF}9.10 & \cellcolor[HTML]{EFEFEF}11.24 & \cellcolor[HTML]{EFEFEF}7.70 & \cellcolor[HTML]{EFEFEF}9.57 & \cellcolor[HTML]{EFEFEF}8.43 & \cellcolor[HTML]{EFEFEF}11.31 & \cellcolor[HTML]{EFEFEF}7.13 & \cellcolor[HTML]{EFEFEF}9.53 & \cellcolor[HTML]{EFEFEF}11.80 & \cellcolor[HTML]{EFEFEF}17.61 & \cellcolor[HTML]{EFEFEF}11.56 & \cellcolor[HTML]{EFEFEF}17.32 \\
w4s50\% & \multirow{-4}{*}{\cellcolor[HTML]{EFEFEF}GQSA} & \cellcolor[HTML]{EFEFEF}11.33 & \cellcolor[HTML]{EFEFEF}\textbf{12.03} & \cellcolor[HTML]{EFEFEF}9.21 & \cellcolor[HTML]{EFEFEF}\textbf{10.85} & \cellcolor[HTML]{EFEFEF}\textbf{10.64} & \cellcolor[HTML]{EFEFEF}\textbf{12.82} & \cellcolor[HTML]{EFEFEF}\textbf{7.80} & \cellcolor[HTML]{EFEFEF}10.93 & \cellcolor[HTML]{EFEFEF}\textbf{13.81} & \cellcolor[HTML]{EFEFEF}\textbf{20.85} & \cellcolor[HTML]{EFEFEF}\textbf{13.56} & \cellcolor[HTML]{EFEFEF}\textbf{20.43} \\ \bottomrule
\end{tabular}
}
\caption{Wikitext2 and C4 perplexity (↓) for LLaMA-1, LLaMA-2, LLaMA-3 and LLaMA-3.1 models, with a context length of 2048.}
\label{tab:ppl_llama}
\end{table*}

Furthermore, to enhance the efficiency of sparse computing, we introduced Stream-K~\cite{osama2023stream}. As show in Figure~\ref{fig:stream}, the classic Slice-K~\cite{guo2024fast} assigns output tiles independently to thread blocks. Each thread block processes one or more rows of the left operand and one or more columns of the right operand to compute the corresponding output tile by slicing along the internal K dimensions. However, when the weight matrix exhibits high sparsity, the uneven distribution of workloads can result in the "straggler" problem, where small workloads cause inefficiencies. Stream-K addresses this issue by decomposing the workload at a finer granularity, allowing multiple thread blocks to collaborate in computing a single output tile.

\section{Experiments}

\subsection{Experimental Settings}
\noindent \textbf{Models and Tasks.} We selected the LLaMA ~\cite{touvron2023llama}, LLaMA-2~\cite{touvron2023llama2}, LLaMA-3, LLaMA-3.1~\cite{dubey2024llama3} and OPT~\cite{zhang2022opt} models to benchmark our method. Following previous studies, we evaluated the model's language modeling capability on the WikiText2~\cite{merity2016pointer} and C4~\cite{raffel2020exploring} datasets. To assess performance on zero-shot tasks, we selected several mainstream benchmarks, including PIQA~\cite{bisk2020piqa}, ARC~\cite{clark2018think}, HellaSwag~\cite{zellers2019hellaswag}, and Winogrande~\cite{sakaguchi2021winogrande}, and conducted evaluations using lm-eval.

\noindent \textbf{Baselines.} We conducted a comprehensive comparison of our method with several recently published techniques in both structured and semi-structured pruning. Given that our implementation achieved INT4 along with 50\% structured pruning, we also compared our approach with pure INT2 quantization. For structured pruning, we compared our results with LLMPruner~\cite{ma2023llm}, SliceGPT~\cite{ashkboos2024slicegpt} and ShortGPT~\cite{men2024shortgpt}. For semi-structured pruning, we utilized SparseGPT~\cite{frantar2023sparsegpt} and Wanda~\cite{sun2023simple} for comparison. Additionally, we selected OmniQuant~\cite{shao2023omniquant}, QuIP~\cite{chee2024quip}, PB-LLM~\cite{shang2023pb}, GPTQ~\cite{frantar2022gptq}, LeanQuant~\cite{zhang2024leanquant}, and SliM-LLM~\cite{huang2024slim} as benchmarks for W2 quantization.

\noindent \textbf{Implementation Details.} To evaluate the performance of GQSA across various configurations, we implemented sparsity levels of 20\%, 30\%, 40\%, and 50\%, using 4-bit weight-only per-group quantization. To strike a balance between model performance and inference speed, a group size of 16 was selected as the optimal configuration. The AdamW optimizer~\cite{loshchilov2017decoupled} with a learning rate of 1e-5 was employed to optimize both BQPO and E2E-OQP. The optimization data was randomly sampled from the WikiText2 and C4 datasets, consisting of 4,096 samples, each containing 2,048 tokens. BQPO was trained for 5 epochs, while E2E-OQP was trained for 2 epochs.

\subsection{Evaluation on Language Generation Tasks}
To assess the performance of GQSA under extreme compression conditions, we first compared its perplexity against baseline method. As shown in Table~\ref{tab:ppl_llama}, GQSA surpasses the performance of current state-of-the-art weight-only per-group quantization methods, including GPTQ, QuIP, OmniQuant, LeanQuant, under a 50\% structured pruning combined with INT4 quantization. It also surpasses mixed-precision quantization models like PB-LLM and SliM-LLM. Furthermore, GQSA achieves comparable results to 2:4 semi-structured pruning while delivering substantial improvements in compression ratio and speedup. Similar results are presented in Table~\ref{tab:ppl_qwen} for Qwen2.5 and Table~\ref{tab:ppl_opt} for OPT models, where GQSA consistently matches or surpasses baseline methods, even under more stringent compression settings. Furthermore, we observe that existing model compression methods often experience significant performance degradation on the latest large language models (e.g., LLaMA-3 and LLaMA-3.1).  In contrast, GQSA demonstrates robust performance even in scenarios where other methods encounter substantial performance degradation.

\begin{table}[ht]
\centering
\resizebox{\columnwidth}{!}{%
\begin{tabular}{cccccccc}
\toprule
Model & Setting & Method & PIQA & ARC-C & ARC-E & Hellaswag & Winogrande \\ \midrule
 &  & ShortGPT & 60.1 & 31.0 & 41.7 & 44.0 & 60.8 \\
 &  & SliceGPT & 67.5 & 34.5 & 55.6 & 55.1 & 62.9 \\ 
 & \multirow{-3}{*}{25\%} & LLM-Pruner & \textbf{75.7} & \textbf{37.2} & 62.0 & 60.1 & 62.2 \\ \cmidrule(l){2-8} 
 & \cellcolor[HTML]{EFEFEF}W4S30\% & \cellcolor[HTML]{EFEFEF}GQSA & \cellcolor[HTML]{EFEFEF}74.32 & \cellcolor[HTML]{EFEFEF}34.98 & \cellcolor[HTML]{EFEFEF}\textbf{66.04} & \cellcolor[HTML]{EFEFEF}\textbf{64.40} & \cellcolor[HTML]{EFEFEF}\textbf{65.98} \\ \cmidrule(l){2-8} 
 &  & ShortGPT & 50.7 & 27.7 & 25.6 & 30.1 & 50.3 \\
 &  & SliceGPT & 58.5 & 27.3 & 43.5 & 43.6 & 57.9 \\
 & \multirow{-3}{*}{40\%} & LLM-Pruner & 70.7 & \textbf{31.3} & 50.7 & 53.5 & 56.1 \\ \cmidrule(l){2-8} 
\multirow{-8}{*}{LLaMA-2-7B} & \cellcolor[HTML]{EFEFEF}W4S40\% & \cellcolor[HTML]{EFEFEF}GQSA & \cellcolor[HTML]{EFEFEF}\textbf{71.27} & \cellcolor[HTML]{EFEFEF}30.72 & \cellcolor[HTML]{EFEFEF}\textbf{61.32} & \cellcolor[HTML]{EFEFEF}\textbf{58.48} & \cellcolor[HTML]{EFEFEF}\textbf{61.48} \\ \midrule
\multicolumn{1}{l}{} &  & ShortGPT & 73.1 & 41.9 & 60.1 & 60.6 & \textbf{70.5} \\
\multicolumn{1}{l}{} &  & SliceGPT & 69.6 & 40.2 & 61.5 & 59.4 & 67.0 \\
\multicolumn{1}{l}{} & \multirow{-3}{*}{25\%} & LLM-Pruner & \textbf{79.4} & \textbf{43.5} & 67.8 & 65.4 & 63.5 \\ \cmidrule(l){2-8} 
\multicolumn{1}{l}{} & \cellcolor[HTML]{EFEFEF}W4S30\% & \cellcolor[HTML]{EFEFEF}GQSA & \cellcolor[HTML]{EFEFEF}75.68 & \cellcolor[HTML]{EFEFEF}39.85 & \cellcolor[HTML]{EFEFEF}\textbf{71.55} & \cellcolor[HTML]{EFEFEF}\textbf{70.45} & \cellcolor[HTML]{EFEFEF}66.54 \\ \cmidrule(l){2-8} 
\multicolumn{1}{l}{} &  & ShortGPT & 62.4 & 32.2 & 44.8 & 47.8 & 62.8 \\
\multicolumn{1}{l}{} &  & SliceGPT & 59.9 & 29.2 & 44.1 & 49.6 & 61.6 \\
\multicolumn{1}{l}{} & \multirow{-3}{*}{40\%} & LLM-Pruner & 75.3 & 35.4 & 56.3 & 60.2 & 57.8 \\ \cmidrule(l){2-8} 
\multicolumn{1}{l}{\multirow{-8}{*}{LLaMA-2-13B}} & \cellcolor[HTML]{EFEFEF}W4S40\% & \cellcolor[HTML]{EFEFEF}GQSA & \cellcolor[HTML]{EFEFEF}\textbf{75.30} & \cellcolor[HTML]{EFEFEF}\textbf{35.82} & \cellcolor[HTML]{EFEFEF}\textbf{66.50} & \cellcolor[HTML]{EFEFEF}\textbf{65.40} & \cellcolor[HTML]{EFEFEF}\textbf{65.98} \\ \bottomrule
\end{tabular}
}
\caption{Zero-shot performance between LLaMA-2-7B and LLaMA-2-13B models under 25\% and 40\% structured pruning, GQSA with 30\% and 40\% structured pruning along with INT4 quantization.}
\label{tab:struct}
\end{table}

\subsection{Evaluation on Zero-Shot Tasks}
To further validate our model, we conducted a detailed comparison of its zero-shot accuracy against baseline methods. Given the limited data availability from these baselines methods, we selected LLaMA-2-7B and LLaMA-2-13B for the analysis. Table~\ref{tab:struct} compares GQSA with structured pruning, where GQSA achieved substantial performance gains at equivalent or higher pruning rates, with these benefits becoming more pronounced at higher pruning levels. Table~\ref{tab:semo-struc} compares GQSA with semi-structured pruning and W2 weight-only per-group quantization. Compared to W2 per-group quantization, GQSA consistently delivered superior performance improvements at the same compression ratio. Under the conditions of 50\% structured pruning with INT4 quantization, GQSA outperformed OmniQuant W2 per-group quantization, yielding average accuracy gains of 5.4\% for LLaMA-2-7B and 5.7\% for LLaMA-2-13B. Given that GQSA operates in a more challenging compression setting than semi-structured pruning, we compare GQSA W4 40\% with semi-structured pruning. Experimental results reveal that GQSA achieves superior performance even with a compression rate 3$\times$ higher than that of 2:4 pruning. Furthermore, GQSA demonstrates significant advantages in both speed and accuracy compared to 2:4 pruning. Considering its compression efficiency and flexibility, GQSA emerges as the clear superior choice.

\begin{table}[ht]
\centering
\resizebox{\columnwidth}{!}{%
\begin{tabular}{cccccccc}
\toprule
Model & Setting & Method & PIQA & ARC-C & ARC-E & Hellaswag & Winogrande \\ \midrule
 &  & OmniQuant & 64.52 & 26.10 & 44.94 & 49.27 & 54.53 \\
 & \multirow{-2}{*}{W2} & LeanQuant & 65.4 & 24.7 & 44.2 & - & 57.4 \\ \cmidrule(l){2-8} 
 & \cellcolor[HTML]{EFEFEF}W4S50\% & \cellcolor[HTML]{EFEFEF}GQSA & \cellcolor[HTML]{EFEFEF}\textbf{68.01} & \cellcolor[HTML]{EFEFEF}\textbf{29.01} & \cellcolor[HTML]{EFEFEF}\textbf{58.33} & \cellcolor[HTML]{EFEFEF}\textbf{52.72} & \cellcolor[HTML]{EFEFEF}\textbf{58.41} \\ \cmidrule(l){2-8} 
 &  & SparseGPT & 70.13 & 29.35 & 61.14 & 56.89 & \textbf{63.14} \\
 & \multirow{-2}{*}{2:4} & Wanda & 70.12 & 30.55 & 61.32 & 55.34 & 62.83 \\ \cmidrule(l){2-8} 
\multirow{-6}{*}{LLaMA-2-7B} & \cellcolor[HTML]{EFEFEF}W4S40\% & \cellcolor[HTML]{EFEFEF}GQSA & \cellcolor[HTML]{EFEFEF}\textbf{71.27} & \cellcolor[HTML]{EFEFEF}\textbf{30.72} & \cellcolor[HTML]{EFEFEF}\textbf{61.32} & \cellcolor[HTML]{EFEFEF}\textbf{58.48} & \cellcolor[HTML]{EFEFEF}61.48 \\ \midrule
 &  & OmniQuant & 68.06 & 30.03 & 57.07 & 56.56 & 52.95 \\
 & \multirow{-2}{*}{W2} & LeanQuant & 70.6 & 28.2 & 56.7 & - & 60.7 \\ \cmidrule(l){2-8} 
 & \cellcolor[HTML]{EFEFEF}W4S50\% & \cellcolor[HTML]{EFEFEF}GQSA & \cellcolor[HTML]{EFEFEF}\textbf{72.47} & \cellcolor[HTML]{EFEFEF}\textbf{33.28} & \cellcolor[HTML]{EFEFEF}\textbf{63.01} & \cellcolor[HTML]{EFEFEF}\textbf{62.11} & \cellcolor[HTML]{EFEFEF}\textbf{62.28} \\ \cmidrule(l){2-8} 
 &  & SparseGPT & 72.74 & 32.59 & 66.04 & 62.78 & 66.54 \\
 & \multirow{-2}{*}{2:4} & Wanda & 73.72 & 34.39 & 66.33 & 63.12 & \textbf{66.93} \\ \cmidrule(l){2-8} 
\multirow{-6}{*}{LLaMA-2-13B} & \cellcolor[HTML]{EFEFEF}W4S40\% & \cellcolor[HTML]{EFEFEF}GQSA & \cellcolor[HTML]{EFEFEF}\textbf{75.30} & \cellcolor[HTML]{EFEFEF}\textbf{35.82} & \cellcolor[HTML]{EFEFEF}\textbf{66.50} & \cellcolor[HTML]{EFEFEF}\textbf{65.40} & \cellcolor[HTML]{EFEFEF}65.98 \\ \bottomrule
\end{tabular}
}
\caption{Zero-shot performance between LLaMA-2-7B and LLaMA-2-13B under W2 quantization method, 50\% semi-structured pruning, and GQSA with 40\% and 50\% structured pruning along with INT4 quantization.}
\label{tab:semo-struc}
\end{table}

\begin{figure}[ht]
  \centering
  \includegraphics[width=\linewidth]{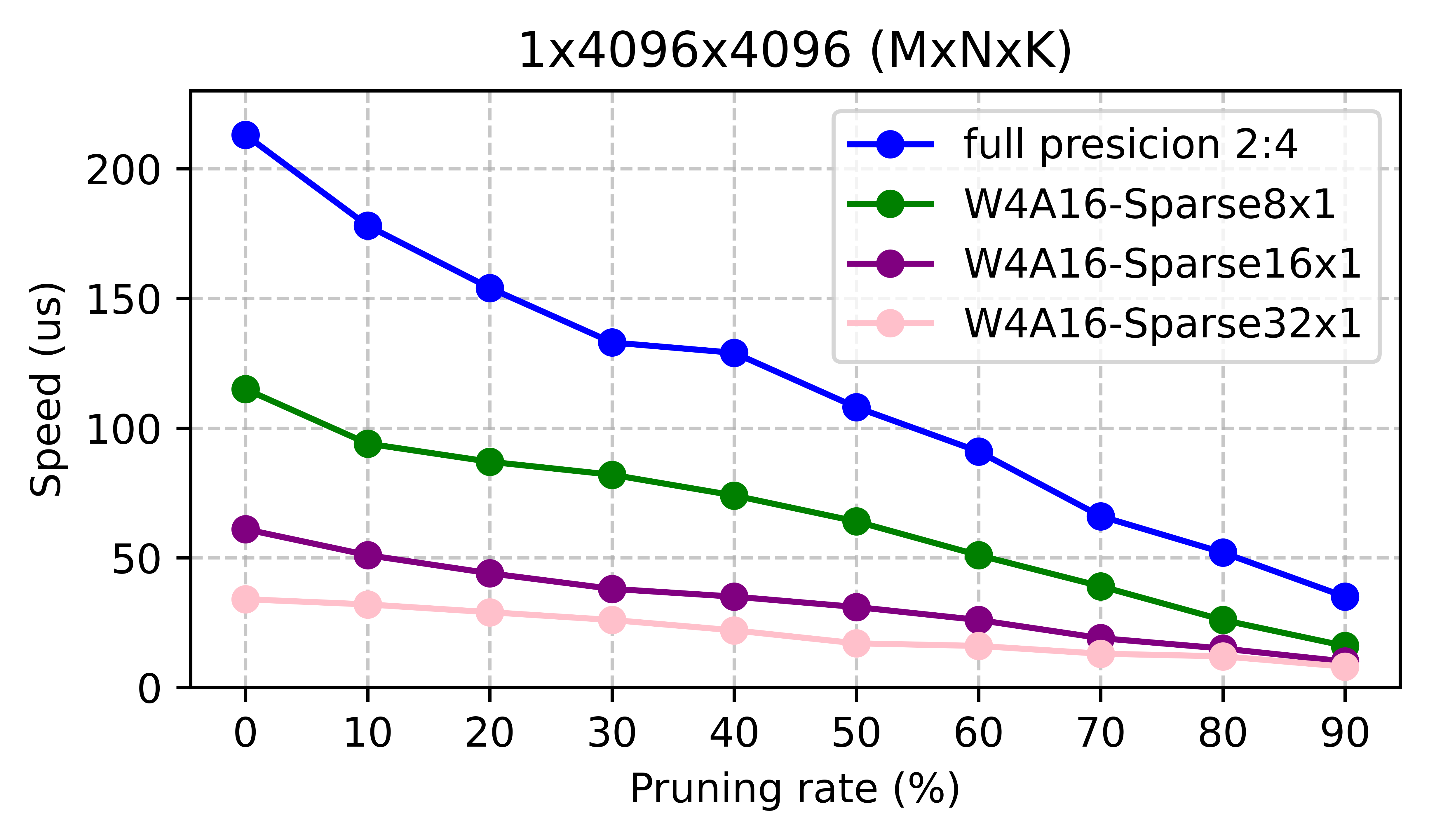}
  \vspace{-0.20in}
  \caption{Comparison of GEMV acceleration of our GQSKernel on RTX 4080.}
  \vspace{-8pt}
  \label{fig:kernel}
\end{figure}

\subsection{Inference Engine Evaluation}
\noindent \textbf{Kernel Benchmark.} We compared GQSKernel with the 2:4 sparse kernel on a (1, 4096) × (4096, 4096) dimension. Due to the flexibility of GQSKernel, it can accommodate varying group sparsity sizes. The experimental results, presented in Figure~\ref{fig:kernel}, show that as sparsity increases, the GEMV computation speed improves. Moreover, GQSKernel consistently outperforms the 2:4 sparse mode across all group granularity settings. At 50\% sparsity, GQSA achieves a 3$\times$ inference speedup compared to the 2:4 sparse mode.

\begin{figure}[ht]
  \centering
  \includegraphics[width=0.9\linewidth]{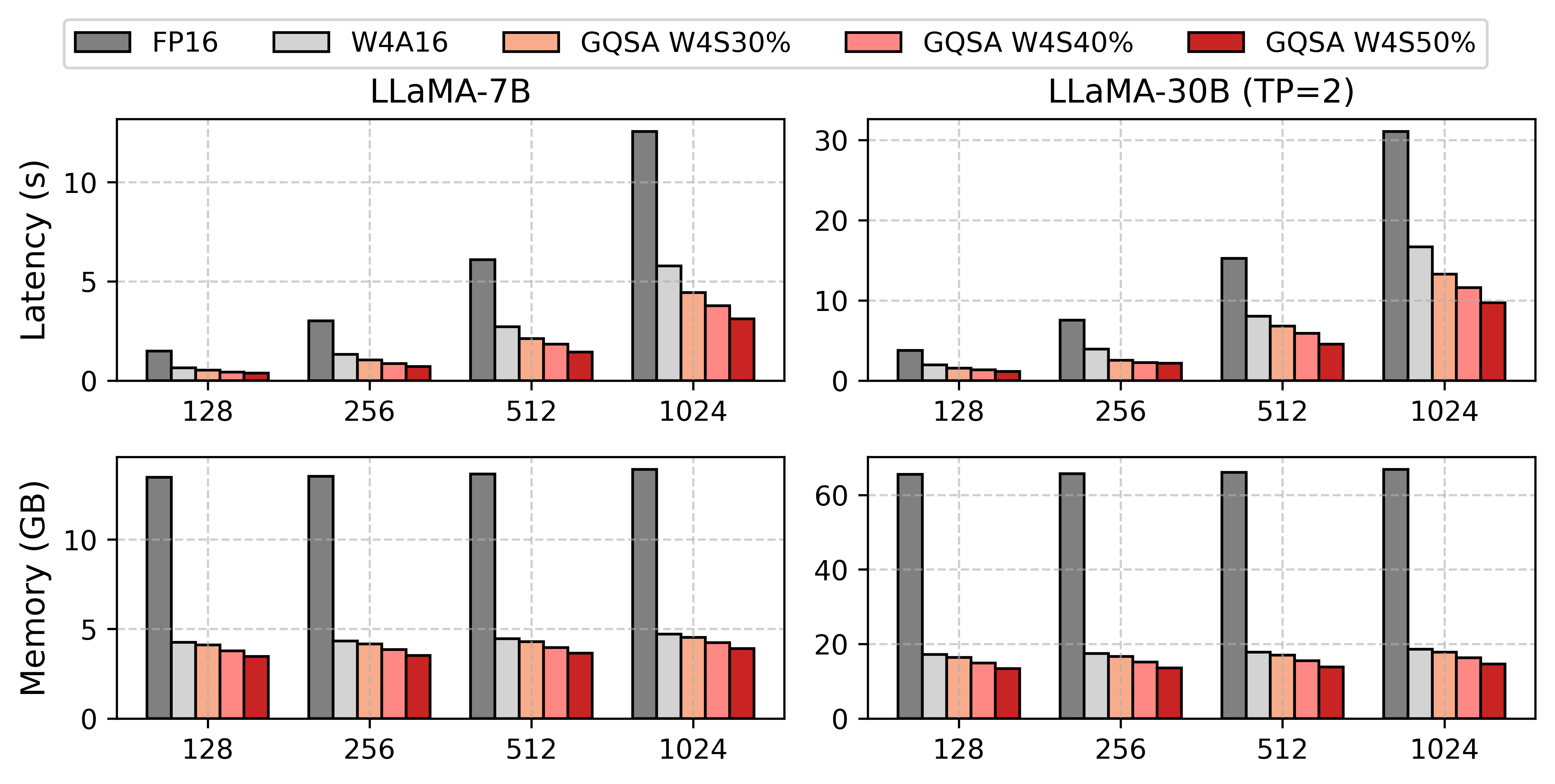}
  \caption{Inference latency (top) and memory usage (bottom)  on an NVIDIA A800-40GB GPU with a fixed input length of 15. W8 results are provided in the Appendix Table~\ref{tab:e2e_speed_appendix}.}
  \vspace{-8pt}
  \label{fig:e2e}
\end{figure}

\noindent \textbf{End-to-end throughput.} 
The acceleration of quantization primarily results from memory access savings, whereas sparsity acceleration arises from both memory access and computational savings. We integrated the GQSKernel into FastTransformer and compared it with the FP16 implementation. The experimental results, as shown in Figure ~\ref{fig:e2e}, indicate that GQSA achieves a 4× reduction in inference latency on the LLaMA-7B model under the GQSA W4S50\% setting with a 1024 output length. Additionally, as presented in Appendix Table~\ref{tab:compare-quantization-pruning}, GQSA further enhances the acceleration potential of the compressed model compared to separate quantization or sparsity methods by simultaneously reducing redundancy in both dimensions of LLMs. For instance, the inference speeds of LLaMA-7B for S50\%, W2, and W4S50\% are 878.90 ms, 475.55 ms, and 377.98 ms, respectively. Overall, GQSA demonstrates the most significant performance improvement.

\begin{table}[ht]
\centering
\resizebox{0.8\columnwidth}{!}{%
\begin{tabular}{cccc}
\toprule
SeqLen & Method & Latency (ms) \\ \midrule
 & W4A16 & 642.24 \\
 & W4 2:4 Pruning & 513.79 \\
\multirow{-3}{*}{128} & \cellcolor[HTML]{EFEFEF}\textbf{GQSA W4 S50\%} & \cellcolor[HTML]{EFEFEF}\textbf{377.98} \\ \midrule
 & W4A16 & 1312.91 \\
 & W4 2:4 Pruning & 1112.96 \\
\multirow{-3}{*}{256} & \cellcolor[HTML]{EFEFEF}\textbf{GQSA W4 S50\%} & \cellcolor[HTML]{EFEFEF}\textbf{699.26} \\ \midrule
 & W4A16 & 2707.26 \\
 & W4 2:4 Pruning & 1966.45 \\
\multirow{-3}{*}{512} & \cellcolor[HTML]{EFEFEF}\textbf{GQSA W4 S50\%} & \cellcolor[HTML]{EFEFEF}\textbf{1433.43} \\ \midrule
 & W4A16 & 5786.8 \\
 & W4 2:4 Pruning & 4118.36 \\
\multirow{-3}{*}{1024} & \cellcolor[HTML]{EFEFEF}\textbf{GQSA W4 S50\%} & \cellcolor[HTML]{EFEFEF}\textbf{3110.54} \\ \bottomrule
\end{tabular}
}
\caption{The inference latency and memory usage of GQSA and 2:4 pruning are compared on an NVIDIA A800-40GB GPU with a fixed input length of 15.}
\label{tab:speed}
\vspace{-8pt}
\end{table}

Additionally, we compared GQSA's performance with that of state-of-the-art sparse schemes, such as SparseGPT and Wanda's 2:4 sparse scheme. The experimental results, presented in Table~\ref{tab:speed}, demonstrate that GQSA outperforms these methods in terms of inference latency and accuracy.

\subsection{Ablation Experiments}
\begin{figure}[th]
  \centering
  \includegraphics[width=0.9\linewidth]{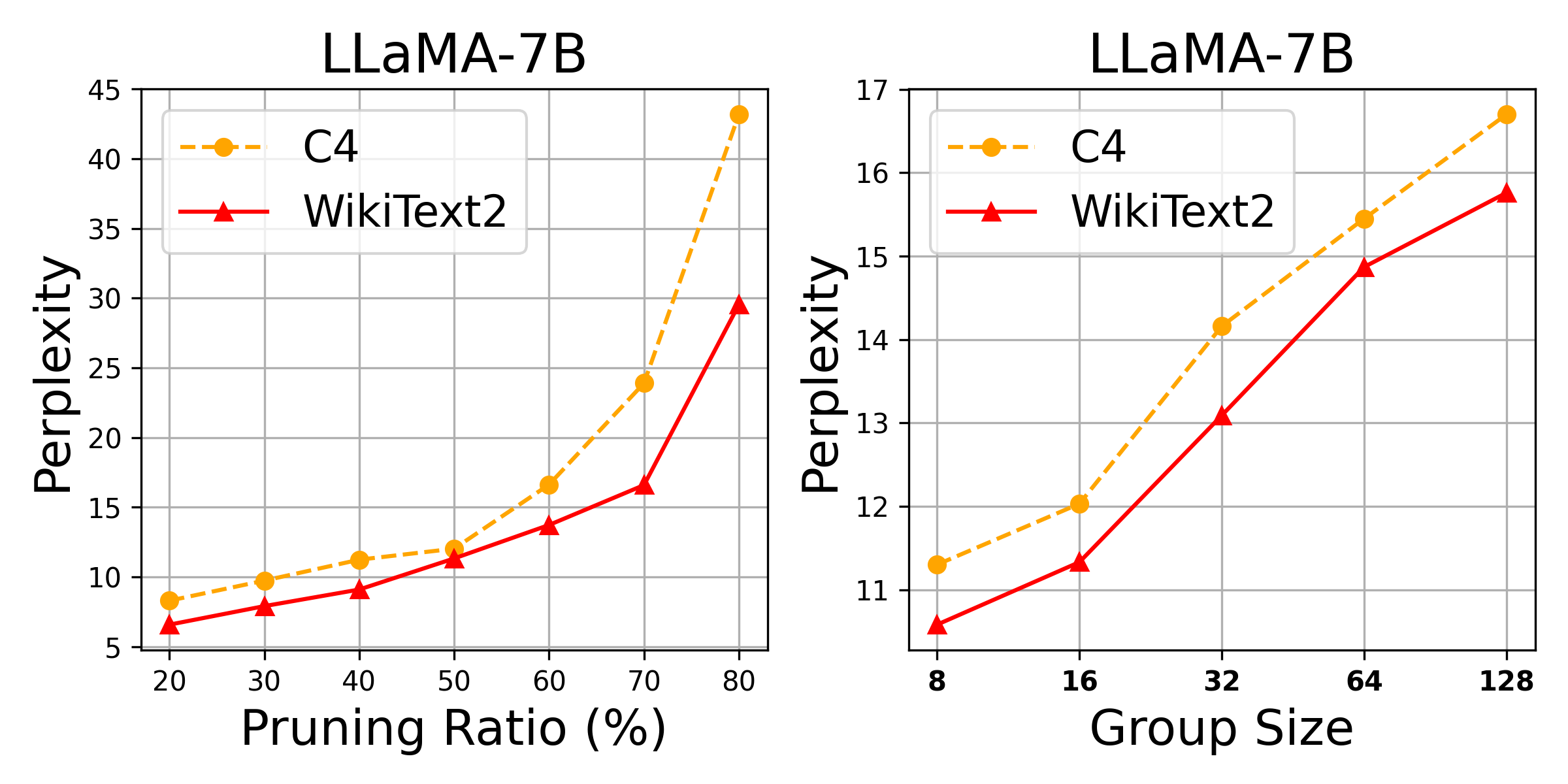}
  \vspace{-0.10in}
  \caption{The ablation studies on the LLaMA-7B model to evaluate the impact of different structured pruning group sizes (right) and sparsity levels (left) on model performance.}
  \vspace{-8pt}
  \label{fig:aba}
\end{figure}

We investigated the impact of group size and sparsity on the performance of the GQSA model. As shown in Figure~\ref{fig:aba} (left), GQSA demonstrates robust performance at sparsity levels of 50\% or lower. When sparsity exceeds 60\%, a noticeable performance degradation occurs. However, even at an extreme sparsity level of 80\%, GQSA achieves a perplexity below 30, avoiding performance collapse. Figure~\ref{fig:aba} (right) illustrates the relationship between group size and model performance. Overall, model performance exhibits a clear correlation with group size. Based on performance considerations, we selected 16 as the default group size for the model.

\section{Conclusion}
We propose GQSA, an efficient sparse acceleration method for the decoding process, compatible with weight-only per-group quantization. Through a comprehensive analysis of LLMs weights, we investigated group sparse modes beyond the 2:4 sparsity mode. To enhance model performance, we implemented a two-stage sparse optimization strategy, comprising BQPO and E2E-OQP. Based on the BSR format, we then developed an efficient sparse inference engine to fully leverage the synergistic benefits of quantization and sparsity. Extensive experimental results demonstrate that GQSA effectively integrates at both the algorithmic and system levels, offering a superior accuracy-speed trade-off compared to traditional 2:4 sparsity and quantization approaches.

\section*{Limitations}
The proposed GQSA extends beyond the 2:4 sparsity pattern to explore group sparsity patterns, enabling efficient compatibility with weight-only per-group quantization. By combining algorithm-level optimizations with a customized inference engine, our approach achieves an improved balance between accuracy and inference speed. However, the current method does not address activation quantization, and due to resource limitations, it has not yet been applied to large language models (LLMs) exceeding 100 billion parameters. These limitations present promising directions for future research, and we are optimistic that they will be addressed in subsequent work.

\section*{Ethics Statement}
This paper introduces a method to tackle the challenges of compressing large language models (LLMs), with the goal of facilitating their wider application and adoption. In the context of current research, ethical considerations surrounding LLMs have received substantial attention. Our findings indicate that the proposed method does not exacerbate existing biases or compromise ethical standards.

\bibliography{custom}

\begin{thebibliography}{40}
\providecommand{\natexlab}[1]{#1}

\bibitem[{Ashkboos et~al.(2024)Ashkboos, Croci, Nascimento, Hoefler, and Hensman}]{ashkboos2024slicegpt}
Saleh Ashkboos, Maximilian~L Croci, Marcelo Gennari~do Nascimento, Torsten Hoefler, and James Hensman. 2024.
\newblock Slicegpt: Compress large language models by deleting rows and columns.
\newblock \emph{arXiv preprint arXiv:2401.15024}.

\bibitem[{Bisk et~al.(2020)Bisk, Zellers, Gao, Choi et~al.}]{bisk2020piqa}
Yonatan Bisk, Rowan Zellers, Jianfeng Gao, Yejin Choi, et~al. 2020.
\newblock Piqa: Reasoning about physical commonsense in natural language.
\newblock In \emph{Proceedings of the AAAI conference on artificial intelligence}, 05, pages 7432--7439.

\bibitem[{Chee et~al.(2024)Chee, Cai, Kuleshov, and De~Sa}]{chee2024quip}
Jerry Chee, Yaohui Cai, Volodymyr Kuleshov, and Christopher~M De~Sa. 2024.
\newblock Quip: 2-bit quantization of large language models with guarantees.
\newblock \emph{Advances in Neural Information Processing Systems}, 36.

\bibitem[{Chen et~al.(2024)Chen, Hu, and Zhang}]{chen2024compressing}
Xiaodong Chen, Yuxuan Hu, and Jing Zhang. 2024.
\newblock Compressing large language models by streamlining the unimportant layer.
\newblock \emph{arXiv preprint arXiv:2403.19135}.

\bibitem[{Clark et~al.(2018)Clark, Cowhey, Etzioni, Khot, Sabharwal, Schoenick, and Tafjord}]{clark2018think}
Peter Clark, Isaac Cowhey, Oren Etzioni, Tushar Khot, Ashish Sabharwal, Carissa Schoenick, and Oyvind Tafjord. 2018.
\newblock Think you have solved question answering? try arc, the ai2 reasoning challenge.
\newblock \emph{arXiv preprint arXiv:1803.05457}.

\bibitem[{Dubey et~al.(2024)Dubey, Jauhri, Pandey, Kadian, Al-Dahle, Letman, Mathur, Schelten, Yang, Fan et~al.}]{dubey2024llama3}
Abhimanyu Dubey, Abhinav Jauhri, Abhinav Pandey, Abhishek Kadian, Ahmad Al-Dahle, Aiesha Letman, Akhil Mathur, Alan Schelten, Amy Yang, Angela Fan, et~al. 2024.
\newblock The llama 3 herd of models.
\newblock \emph{arXiv preprint arXiv:2407.21783}.

\bibitem[{Egiazarian et~al.(2024)Egiazarian, Panferov, Kuznedelev, Frantar, Babenko, and Alistarh}]{egiazarian2024extreme}
Vage Egiazarian, Andrei Panferov, Denis Kuznedelev, Elias Frantar, Artem Babenko, and Dan Alistarh. 2024.
\newblock Extreme compression of large language models via additive quantization.
\newblock \emph{arXiv preprint arXiv:2401.06118}.

\bibitem[{Fang et~al.(2024)Fang, Yin, Muralidharan, Heinrich, Pool, Kautz, Molchanov, and Wang}]{fang2024maskllm}
Gongfan Fang, Hongxu Yin, Saurav Muralidharan, Greg Heinrich, Jeff Pool, Jan Kautz, Pavlo Molchanov, and Xinchao Wang. 2024.
\newblock Maskllm: Learnable semi-structured sparsity for large language models.
\newblock \emph{arXiv preprint arXiv:2409.17481}.

\bibitem[{Frantar and Alistarh(2023)}]{frantar2023sparsegpt}
Elias Frantar and Dan Alistarh. 2023.
\newblock Sparsegpt: Massive language models can be accurately pruned in one-shot.
\newblock In \emph{International Conference on Machine Learning}, pages 10323--10337. PMLR.

\bibitem[{Frantar et~al.(2022)Frantar, Ashkboos, Hoefler, and Alistarh}]{frantar2022gptq}
Elias Frantar, Saleh Ashkboos, Torsten Hoefler, and Dan Alistarh. 2022.
\newblock Gptq: Accurate post-training quantization for generative pre-trained transformers.
\newblock \emph{arXiv preprint arXiv:2210.17323}.

\bibitem[{Gerganov(2024)}]{gerganov2024llama}
M.~Gerganov. 2024.
\newblock \href {https://github.com/ggerganov/llama.cpp} {llama.cpp: A high-performance implementation of llama}.
\newblock Accessed: 2024-12-12.

\bibitem[{Gu et~al.(2024)Gu, Dong, Wei, and Huang}]{gu2024minillm}
Yuxian Gu, Li~Dong, Furu Wei, and Minlie Huang. 2024.
\newblock Minillm: Knowledge distillation of large language models.
\newblock In \emph{The Twelfth International Conference on Learning Representations}.

\bibitem[{Guo et~al.(2024)Guo, Brandon, Cholakov, Ragan-Kelley, Xing, and Kim}]{guo2024fast}
Han Guo, William Brandon, Radostin Cholakov, Jonathan Ragan-Kelley, Eric~P Xing, and Yoon Kim. 2024.
\newblock Fast matrix multiplications for lookup table-quantized llms.
\newblock \emph{arXiv preprint arXiv:2407.10960}.

\bibitem[{Han et~al.(2016)Han, Mao, and Dally}]{han2015deep_compression}
Song Han, Huizi Mao, and William~J Dally. 2016.
\newblock Deep compression: Compressing deep neural networks with pruning, trained quantization and huffman coding.
\newblock \emph{International Conference on Learning Representations (ICLR)}.

\bibitem[{Han et~al.(2015)Han, Pool, Tran, and Dally}]{han2015learning}
Song Han, Jeff Pool, John Tran, and William Dally. 2015.
\newblock Learning both weights and connections for efficient neural network.
\newblock \emph{Advances in neural information processing systems}, 28.

\bibitem[{Huang et~al.(2024)Huang, Qin, Liu, Li, Liu, Benini, Magno, and Qi}]{huang2024slim}
Wei Huang, Haotong Qin, Yangdong Liu, Yawei Li, Xianglong Liu, Luca Benini, Michele Magno, and Xiaojuan Qi. 2024.
\newblock Slim-llm: Salience-driven mixed-precision quantization for large language models.
\newblock \emph{arXiv preprint arXiv:2405.14917}.

\bibitem[{Lee et~al.(2024)Lee, Jin, Kim, Kim, and Park}]{lee2024owq}
Changhun Lee, Jungyu Jin, Taesu Kim, Hyungjun Kim, and Eunhyeok Park. 2024.
\newblock Owq: Outlier-aware weight quantization for efficient fine-tuning and inference of large language models.
\newblock In \emph{Proceedings of the AAAI Conference on Artificial Intelligence}, volume~38, pages 13355--13364.

\bibitem[{Lin et~al.(2024)Lin, Tang, Tang, Yang, Chen, Wang, Xiao, Dang, Gan, and Han}]{lin2024awq}
Ji~Lin, Jiaming Tang, Haotian Tang, Shang Yang, Wei-Ming Chen, Wei-Chen Wang, Guangxuan Xiao, Xingyu Dang, Chuang Gan, and Song Han. 2024.
\newblock Awq: Activation-aware weight quantization for on-device llm compression and acceleration.
\newblock \emph{Proceedings of Machine Learning and Systems}, 6:87--100.

\bibitem[{Liu et~al.(2022)Liu, Tao, Feng, and Zhao}]{liu2022multi}
Chang Liu, Chongyang Tao, Jiazhan Feng, and Dongyan Zhao. 2022.
\newblock Multi-granularity structural knowledge distillation for language model compression.
\newblock In \emph{Proceedings of the 60th Annual Meeting of the Association for Computational Linguistics (Volume 1: Long Papers)}, pages 1001--1011.

\bibitem[{Loshchilov(2017)}]{loshchilov2017decoupled}
I~Loshchilov. 2017.
\newblock Decoupled weight decay regularization.
\newblock \emph{arXiv preprint arXiv:1711.05101}.

\bibitem[{Ma et~al.(2023)Ma, Fang, and Wang}]{ma2023llm}
Xinyin Ma, Gongfan Fang, and Xinchao Wang. 2023.
\newblock Llm-pruner: On the structural pruning of large language models.
\newblock \emph{Advances in neural information processing systems}, 36:21702--21720.

\bibitem[{Men et~al.(2024)Men, Xu, Zhang, Wang, Lin, Lu, Han, and Chen}]{men2024shortgpt}
Xin Men, Mingyu Xu, Qingyu Zhang, Bingning Wang, Hongyu Lin, Yaojie Lu, Xianpei Han, and Weipeng Chen. 2024.
\newblock Shortgpt: Layers in large language models are more redundant than you expect.
\newblock \emph{arXiv preprint arXiv:2403.03853}.

\bibitem[{Merity et~al.(2016)Merity, Xiong, Bradbury, and Socher}]{merity2016pointer}
Stephen Merity, Caiming Xiong, James Bradbury, and Richard Socher. 2016.
\newblock Pointer sentinel mixture models.
\newblock \emph{arXiv preprint arXiv:1609.07843}.

\bibitem[{Mishra et~al.(2021)Mishra, Latorre, Pool, Stosic, Stosic, Venkatesh, Yu, and Micikevicius}]{mishra2021accelerating}
Asit Mishra, Jorge~Albericio Latorre, Jeff Pool, Darko Stosic, Dusan Stosic, Ganesh Venkatesh, Chong Yu, and Paulius Micikevicius. 2021.
\newblock Accelerating sparse deep neural networks.
\newblock \emph{arXiv preprint arXiv:2104.08378}.

\bibitem[{Mozaffari and Dehnavi(2024)}]{mozaffari2024slim}
Mohammad Mozaffari and Maryam~Mehri Dehnavi. 2024.
\newblock Slim: One-shot quantized sparse plus low-rank approximation of llms.
\newblock \emph{arXiv preprint arXiv:2410.09615}.

\bibitem[{Osama et~al.(2023)Osama, Merrill, Cecka, Garland, and Owens}]{osama2023stream}
Muhammad Osama, Duane Merrill, Cris Cecka, Michael Garland, and John~D Owens. 2023.
\newblock Stream-k: Work-centric parallel decomposition for dense matrix-matrix multiplication on the gpu.
\newblock In \emph{Proceedings of the 28th ACM SIGPLAN Annual Symposium on Principles and Practice of Parallel Programming}, pages 429--431.

\bibitem[{Raffel et~al.(2020)Raffel, Shazeer, Roberts, Lee, Narang, Matena, Zhou, Li, and Liu}]{raffel2020exploring}
Colin Raffel, Noam Shazeer, Adam Roberts, Katherine Lee, Sharan Narang, Michael Matena, Yanqi Zhou, Wei Li, and Peter~J Liu. 2020.
\newblock Exploring the limits of transfer learning with a unified text-to-text transformer.
\newblock \emph{Journal of machine learning research}, 21(140):1--67.

\bibitem[{Sakaguchi et~al.(2021)Sakaguchi, Bras, Bhagavatula, and Choi}]{sakaguchi2021winogrande}
Keisuke Sakaguchi, Ronan~Le Bras, Chandra Bhagavatula, and Yejin Choi. 2021.
\newblock Winogrande: An adversarial winograd schema challenge at scale.
\newblock \emph{Communications of the ACM}, 64(9):99--106.

\bibitem[{Shang et~al.(2023)Shang, Yuan, Wu, and Dong}]{shang2023pb}
Yuzhang Shang, Zhihang Yuan, Qiang Wu, and Zhen Dong. 2023.
\newblock Pb-llm: Partially binarized large language models.
\newblock \emph{arXiv preprint arXiv:2310.00034}.

\bibitem[{Shao et~al.(2023)Shao, Chen, Zhang, Xu, Zhao, Li, Zhang, Gao, Qiao, and Luo}]{shao2023omniquant}
Wenqi Shao, Mengzhao Chen, Zhaoyang Zhang, Peng Xu, Lirui Zhao, Zhiqian Li, Kaipeng Zhang, Peng Gao, Yu~Qiao, and Ping Luo. 2023.
\newblock Omniquant: Omnidirectionally calibrated quantization for large language models.
\newblock \emph{arXiv preprint arXiv:2308.13137}.

\bibitem[{Sun et~al.(2023)Sun, Liu, Bair, and Kolter}]{sun2023simple}
Mingjie Sun, Zhuang Liu, Anna Bair, and J~Zico Kolter. 2023.
\newblock A simple and effective pruning approach for large language models.
\newblock \emph{arXiv preprint arXiv:2306.11695}.

\bibitem[{Touvron et~al.(2023{\natexlab{a}})Touvron, Lavril, Izacard, Martinet, Lachaux, Lacroix, Rozi{\`e}re, Goyal, Hambro, Azhar et~al.}]{touvron2023llama}
Hugo Touvron, Thibaut Lavril, Gautier Izacard, Xavier Martinet, Marie-Anne Lachaux, Timoth{\'e}e Lacroix, Baptiste Rozi{\`e}re, Naman Goyal, Eric Hambro, Faisal Azhar, et~al. 2023{\natexlab{a}}.
\newblock Llama: Open and efficient foundation language models.
\newblock \emph{arXiv preprint arXiv:2302.13971}.

\bibitem[{Touvron et~al.(2023{\natexlab{b}})Touvron, Martin, Stone, Albert, Almahairi, Babaei, Bashlykov, Batra, Bhargava, Bhosale et~al.}]{touvron2023llama2}
Hugo Touvron, Louis Martin, Kevin Stone, Peter Albert, Amjad Almahairi, Yasmine Babaei, Nikolay Bashlykov, Soumya Batra, Prajjwal Bhargava, Shruti Bhosale, et~al. 2023{\natexlab{b}}.
\newblock Llama 2: Open foundation and fine-tuned chat models.
\newblock \emph{arXiv preprint arXiv:2307.09288}.

\bibitem[{Tseng et~al.(2024)Tseng, Chee, Sun, Kuleshov, and De~Sa}]{tseng2024quip}
Albert Tseng, Jerry Chee, Qingyao Sun, Volodymyr Kuleshov, and Christopher De~Sa. 2024.
\newblock Quip\#: Even better llm quantization with hadamard incoherence and lattice codebooks.
\newblock \emph{arXiv preprint arXiv:2402.04396}.

\bibitem[{Wang et~al.(2025)Wang, Mao, Tang, Du, Guan, and Xue}]{wang2025compression}
Weilan Wang, Yu~Mao, Dongdong Tang, Hongchao Du, Nan Guan, and Chun~Jason Xue. 2025.
\newblock When compression meets model compression: Memory-efficient double compression for large language models.
\newblock \emph{arXiv preprint arXiv:2502.15443}.

\bibitem[{Wang et~al.(2024)Wang, Zhang, Zhao, Farnia, and Yu}]{wang2024moreaupruner}
Zixiao Wang, Jingwei Zhang, Wenqian Zhao, Farzan Farnia, and Bei Yu. 2024.
\newblock Moreaupruner: Robust pruning of large language models against weight perturbations.
\newblock \emph{arXiv preprint arXiv:2406.07017}.

\bibitem[{Zellers et~al.(2019)Zellers, Holtzman, Bisk, Farhadi, and Choi}]{zellers2019hellaswag}
Rowan Zellers, Ari Holtzman, Yonatan Bisk, Ali Farhadi, and Yejin Choi. 2019.
\newblock Hellaswag: Can a machine really finish your sentence?
\newblock In \emph{Proceedings of the 57th Annual Meeting of the Association for Computational Linguistics}.

\bibitem[{Zeng et~al.(2024)Zeng, Liu, Xie, Liu, Wang, Wei, Yang, Chen, and Mei}]{zeng2024abq}
Chao Zeng, Songwei Liu, Yusheng Xie, Hong Liu, Xiaojian Wang, Miao Wei, Shu Yang, Fangmin Chen, and Xing Mei. 2024.
\newblock Abq-llm: Arbitrary-bit quantized inference acceleration for large language models.
\newblock \emph{arXiv preprint arXiv:2408.08554}.

\bibitem[{Zhang et~al.(2022)Zhang, Roller, Goyal, Artetxe, Chen, Chen, Dewan, Diab, Li, Lin et~al.}]{zhang2022opt}
Susan Zhang, Stephen Roller, Naman Goyal, Mikel Artetxe, Moya Chen, Shuohui Chen, Christopher Dewan, Mona Diab, Xian Li, Xi~Victoria Lin, et~al. 2022.
\newblock Opt: Open pre-trained transformer language models.
\newblock \emph{arXiv preprint arXiv:2205.01068}.

\bibitem[{Zhang and Shrivastava(2024)}]{zhang2024leanquant}
Tianyi Zhang and Anshumali Shrivastava. 2024.
\newblock Leanquant: Accurate large language model quantization with loss-error-aware grid.
\newblock \emph{arXiv preprint arXiv:2407.10032}.

\end{thebibliography}

\newpage
\appendix
\section*{Appendix}
\label{sec:appendix}

\setcounter{section}{0}
\renewcommand{\thesection}{\Alph{section}} 

\section{Training Efficiency of GQSA} \label{ap:train_efficien}
Table~\ref{tab:memory} lists the memory and time required to train the Lllama-2 model using GQSA. The results show that GQSA requires only minimal resource overhead, and the 7B model only takes less than 10 hours to train with 9.3GB of memory, which is much less than the 14GB  memory requirement to load the FP16 model, which is very efficient. It is also significantly better than other pruning methods, such as LLM-Pruner, which requires 18GB and takes less training time.

\begin{table}[ht]
\centering
\resizebox{\columnwidth}{!}{%
\begin{tabular}{ccccc}
\toprule
\multirow{2}{*}{LLaMA-2} & \multicolumn{2}{c}{BQPO} & \multicolumn{2}{c}{E2E-OQP} \\ \cmidrule(l){2-5} 
                       & Memory (GB)  & Time (h)  & Memory (GB)    & Time (h)   \\ \midrule
7B             & 9.3          & 5.1       & 7.6              & 4.2        \\
13B            & 14.3         & 7.3       & 11.7           & 6.4        \\ \bottomrule
\end{tabular}
}
\caption{Detailed training time and training memory for GQSA at different model sizes and quantization bits on a single A100-40GB GPU.}
\label{tab:memory}
\end{table}

\section{The effects of BQPO and E2E-OQP on the model’s performance}

Table~\ref{tab:aba-acc} presents the impact of BQPO and E2E-OQP on model performance. BQPO optimizes weights in a block-wise manner, effectively preserving the performance of the GQS model. Finally, E2E-OQP, which accounts for cross-layer errors, yields the best model performance.

\begin{table}[ht]
\centering
\resizebox{\columnwidth}{!}{%
\begin{tabular}{ccccc}
\toprule
\multirow{2}{*}{Method} & \multicolumn{2}{c}{LLaMA-13B} & \multicolumn{2}{c}{LLaMA-2-13B} \\ \cmidrule(l){2-5} 
 & WikiText2 & C4 & WikiText2 & C4 \\ \midrule
BQPO & 12.90 & 13.39 & 10.55 & 13.56 \\
BQPO+E2E-OQP & 9.21 & 10.85 & 7.80 & 10.93 \\ \bottomrule
\end{tabular}
}
\caption{The effectiveness of BQPO and E2E-OQP methods for compressing LLaMA-13B and LLaMA-2-13B models.}
\label{tab:aba-acc}
\end{table}

\section{GQSA performance under weight-activation quantization}

Unlike other algorithms that are limited to weight-only quantization, weight-activation quantization or model pruning, GQSA can not only efficiently combine pruning with weight-only quantization, but also support pruning with weight activation quantization. Our GPU-friendly grouped semi-structured sparse solution can be seamlessly combined with weight-only quantization or weight+activation quantization. On the basis of quantization, we can further improve performance by skipping some operations through sparsity.

\begin{table}[ht]
\centering
\resizebox{\columnwidth}{!}{%
\begin{tabular}{cccc}
\toprule
Model       & Settings  & WikiText2 & C4    \\ \midrule
LLaMA-2-7B  & W4A8S50\% & 7.84      & 11.04 \\
LLaMA-2-13B & W4A8S50\% & 14.09     & 21.26 \\ \bottomrule
\end{tabular}
}
\caption{Performance comparison of GQSA with weight-activation quantization.}
\label{tab:gqsa-activation}
\end{table}

As show in Table~\ref{tab:gqsa-activation}, GQSA effectively preserves model accuracy under W4A8S50\% quantization for both LLaMA-2-7B and LLaMA-2-13B architectures, maintaining strong performance despite simultaneous weight and activation quantization with 50\% sparsity.

\section{Comparison of Quantized \& Pruned Works}

\textbf{A comparison with SparseGPT’s joint sparsification and quantization.} In SparseGPT’s report, "Joint Sparsification \& Quantization" performs worse than "Sparsification-only," so we initially did not include it in our main content. However, for completeness, the following Table~\ref{tab:gqsa-sparsegpt} presents a direct comparison between GQSA and SparseGPT's "Joint Sparsification \& Quantization" on LLaMA-2-13B and LLaMA-3-8B. The results demonstrate that GQSA provides a more significant performance advantage.

\begin{table}[ht]
\centering
\resizebox{\columnwidth}{!}{%
\begin{tabular}{ccccc}
\toprule
\multirow{2}{*}{Method} & \multicolumn{2}{c}{LlaMA-2-13B} & \multicolumn{2}{c}{LLaMA-3-8B} \\ \cmidrule(l){2-5} 
 & WikiText2 & C4 & WikiText2 & C4 \\ \midrule
SparseGPT 2:4 & 8.32 & 11.30 & 16.56 & 22.99 \\
SparseGPT 2:4+INT4 & 9.25 & 12.74 & 19.43 & 26.34 \\
GQSA W4S50\% & 7.80 & 10.93 & 13.84 & 20.85 \\ \bottomrule
\end{tabular}
}
\caption{Performance comparison of GQSA with SparseGPT.}
\label{tab:gqsa-sparsegpt}
\end{table}

\noindent \textbf{Comparison with contemporaneous works.} To further validate the superiority of GQSA, we conduct comparative evaluations with contemporaneous methods including SliM-LoRA~\cite{mozaffari2024slim} and DC-W8A8~\cite{wang2025compression}. \textbf{SliM-LoRA} employs 4-bit weight quantization combined with Wanda's 2:4 pruning but fails to overcome the limitations of semi-structured sparsity. Since NVIDIA's 2:4 Tensor Cores do not support weight-only quantization, the inference acceleration benefits remain limited. Additionally, the 2:4 sparse format retains randomly positioned neurons and requires storing an equal amount of metadata to record their locations, preventing effective memory compression. SliM-LoRA also introduces the LoRA-Adapter; however, due to the quantization and sparsity of the main network, the LoRA-Adapter cannot be directly integrated and must be stored separately, increasing inference complexity. According to the SliM paper, its sparse quantization matrix can even reduce inference speed on the A100 GPU. In contrast, GQSA's sparse quantization achieves a 4.3 $\times$
 inference speedup over FP16, highlighting SliM’s shortcomings in both inference acceleration and memory compression. \textbf{DC-W8A8} incorporates sparsity into W8A8 quantization but relies on unstructured sparsity with a sparsity rate of only 20\%, offering minimal memory compression benefits. As stated in its paper, DC-W8A8 achieves only a 2.2$\times$ compression ratio compared to FP16, whereas GQSA achieves a 4.3$\times$ compression ratio. Moreover, GQSA significantly outperforms DC-W8A8 in inference acceleration.

\begin{table}[ht]
\centering
\resizebox{\columnwidth}{!}{%
\begin{tabular}{ccccc}
\toprule
\multirow{2}{*}{Method} & \multicolumn{2}{c}{OPT} & \multicolumn{2}{c}{LLaMA-2} \\ \cmidrule(l){2-5} 
 & 6.7B & 13B & 7B & 13B \\ \midrule
SliM-LoRA & 47.08 & 47.96 & 54.26 & 57.85 \\
DC-W8A8 & 48.55 & - & \textbf{60.89} & - \\
GQSA W4S50\% & \textbf{53.26} & \textbf{56.39} & 59.36 & \textbf{64.96} \\ \bottomrule
\end{tabular}
}
\caption{Performance comparison between GQSA and contemporaneous methods.}
\label{tab:gqsa-slim-lora}
\end{table}

Table~\ref{tab:gqsa-slim-lora} presents the comparative evaluation of average accuracy on zero-shot tasks across different methods. The experimental results demonstrate that GQSA consistently outperforms both SliM and DC-W8A8 in terms of overall performance. Furthermore, GQSA achieves superior acceleration ratios and compression rates, while maintaining competitive accuracy. These advantages make GQSA particularly suitable for edge-side inference scenarios, where both computational efficiency and model compactness are critical.

\section{A comparison of the effects of pruning and quantization on inference performance}

\begin{table}[ht]
\centering
\resizebox{\columnwidth}{!}{%
\begin{tabular}{cccc}
\toprule
Setting          & WikiText2      & C4             & Inference speed (ms) \\ \midrule
0\%              & 5.47           & 6.97           & 1490.50              \\
S20\%            & 7.67           & 9.10           & 1370.35              \\
S30\%            & 9.34           & 11.27          & 1181.25              \\
S40\%            & 10.84          & 16.38          & 1035.15              \\
S50\%            & 14.56          & 21.09          & 878.90               \\
S60\%            & 25.76          & 37.49          & 671.98               \\ \midrule
W8               & 5.50           & 7.01           & 868.35               \\
W4               & 5.72           & 7.25           & 642.24               \\
W2               & 36.43          & 40.34          & 475.55               \\ \midrule
\textbf{W4S50\%} & \textbf{10.64} & \textbf{12.82} & \textbf{377.98}               \\ \bottomrule
\end{tabular}
}
\caption{Performance comparison of GQSA with naive pruning and naive quantization in the extreme compression setting on LLaMA-2-7B.}
\label{tab:compare-quantization-pruning}
\end{table}

As demonstrated in Table~\ref{tab:compare-quantization-pruning}, we will highlight the comprehensive performance advantages of GQSA over single pruning and quantization methods from two perspectives.

\noindent \textbf{From the Perspective of Algorithm Accuracy:} Both quantization and sparsity, when applied individually, can lead to significant accuracy degradation under extreme compression settings. For instance, the PPL test results under S60\% and W2 configurations demonstrate considerable performance loss. However, combining these two strategies allows for higher compression rates while better preserving model performance compared to using either strategy alone. As an example, using the LLaMA-2-7B WikiText2 benchmark, the results for W2, S60\%, and W4S50\% are 36.44, 25.76, and 10.64, respectively.

\noindent \textbf{From the Perspective of Inference Speed:} The acceleration benefit of quantization primarily arises from reduced memory access, while the acceleration benefit of sparsity stems from both memory and computational savings. For pure quantization or pure sparsity, the acceleration benefit diminishes as the compression rate increases. GQSA, however, enhances the upper limit of the acceleration benefit by simultaneously reducing redundancy in both dimensions (quantization and sparsity). For example, in the case of LLaMA-2-7B, the inference speeds for S60\%, W2, and W4S50\% are 671.98, 475.55, and 377.98, respectively.

\section{Combining the advantages of structured pruning and group quantization}

As show in Table~\ref{tab:gqsa-single-quantization} the acceleration benefits of quantization primarily stem from reduced memory access, while sparsity accelerates inference by saving both memory and computation (as sparse groups do not need to be stored, read, or computed). When applying only the quantization strategy, the LLM's acceleration benefit does not increase exponentially as the bit-width of $W$ decreases. Instead, it faces diminishing returns, as the performance bottleneck shifts from memory access to computation as the quantization bit-width is reduced. In contrast, GQSA can further accelerate deep quantization models by skipping redundant calculations, thereby pushing the upper limit of acceleration benefits. For example, in the case of LLaMA-2-7B, the measured inference speed is 20\% faster with W4S50 than with W2.

\begin{table}[th]
\centering
\resizebox{\columnwidth}{!}{%
\begin{tabular}{ccc}
\toprule
Model                       & Setting          & Inference speed (ms) \\ \midrule
\multirow{3}{*}{LLaMA-2-7B} & W4               & 642.24              \\
                            & W2               & 475.55              \\
                            & \textbf{W4S50\%} & \textbf{377.98}     \\ \bottomrule
\end{tabular}
}
\caption{Comparison of inference speed between GQSA and single quantization.}
\label{tab:gqsa-single-quantization}
\end{table}

\section{Comparison of GQSA with Vector Quantization}

Some of the latest low-bit quantization methods, such as AQLM~\cite{egiazarian2024extreme} and QuIP\#~\cite{tseng2024quip}, employ vector quantization (VQ), which differs from uniform quantization techniques like GQSA. VQ constructs codebooks by learning the underlying data distribution, enabling better data preservation and potentially higher model performance. However, VQ methods rely on pre-trained codebooks (e.g., the E8P codebook used in QuIP\# and the multi-codebook scheme in AQLM), which introduce considerable computational overhead during both training and inference. This makes them less practical for real-world deployment.

\begin{table}[ht]
\centering
\resizebox{\columnwidth}{!}{%
\begin{tabular}{cccc}
\hline
Method & WikiText2 & C4 & Tokens Per Second \\ \hline
QuIP\# W2 & 6.06 & 8.07 & 71.09 \\
AQLM W2 & 5.60 & 7.47 & 68.1 \\
GQSA W4S50\% & 7.80 & 10.93 & 228.95 \\ \hline
\end{tabular}
}
\caption{Comparison between GQSA and Vector Quantization}
\label{tab:gqsa-vector}
\end{table}

In contrast, GQSA combines uniform quantization with high sparsity to enable efficient inference acceleration in practical scenarios. As show in Table~\ref{tab:gqsa-vector}, while it may slightly underperform VQ-based methods like QuIP\# and AQLM in terms of accuracy, it significantly outpaces them in inference speed—achieving up to 3.3$\times$ the speed of vector quantization methods under a small accuracy trade-off. No single method perfectly balances accuracy and computational efficiency; GQSA prioritizes inference speed, accepting a minor compromise in model accuracy to achieve substantial gains in performance.

\section{Inference throughput of GQSA}
As show in Table~\ref{tab:throughput}, we evaluated the throughput of the GQSA model based on FastTransformer on an Nvidia A100 80 GB GPU. The results demonstrate that, compared to the pure W8 and W4 configurations, GQSA's W8S50\% and W4S50\% configurations achieved a 60\% improvement in throughput.
\begin{table}[t]
\centering
\resizebox{\columnwidth}{!}{%
\begin{tabular}{ccc}
\toprule
Setting        & LLaMA-7B        & LLaMA-13B       \\ \midrule
FP             & 92.69           & 50.68           \\
W8             & 156.40          & 95.78           \\
\textbf{W8S50} & \textbf{263.64} & \textbf{158.99} \\
W4             & 202.81          & 137.92          \\
\textbf{W4S50} & \textbf{343.43} & \textbf{228.95} \\ \bottomrule
\end{tabular}
}
\caption{Inference throughput (tokens per second) of GQSA on the NVIDIA A100 80GB.}
\label{tab:throughput}
\end{table}

\section{Differences from Sparse Methods in Traditional CNNs}
Although previous work, such as PatDNN, introduced semi-structured sparsity in CNN networks, we believe our work contributes to the field in two core aspects: \textbf{First}, we have significantly advanced the engineering implementation of semi-structured sparsity. Notably, we introduced the "task-centric" parallel strategy, replacing the widely-used "data-centric" parallel approach in the industry. This shift effectively addresses the issue of unbalanced load across computing units, resulting in a substantial speedup of 1.3$\times$ to 1.5$\times$ for individual operators, thus achieving a new state-of-the-art in engineering performance. \textbf{Second}, while the GEMM operator in traditional CNN networks typically adopts the "N$\times$1" sparse mode, we propose the "1$\times$N" sparse mode tailored to the characteristics of LLM models. This innovation better preserves outliers within the channel and is fundamentally different from the traditional "N$\times$1" mode in terms of engineering implementation.

We believe innovation is not solely about proposing "new concepts" or "new strategies" but also about selecting the most appropriate approaches to address real technical challenges and pushing the performance boundaries. Currently, the LLM field faces significant inference cost challenges, and relying exclusively on quantization techniques has nearly reached its performance optimization limits. Our work contributes to further enhancing performance based on quantization models and has led to a SOTA breakthrough in semi-structured sparsity technology within the LLM field. The pursuit of higher performance limits and greater industrial applicability reflects a key aspect of innovation.

\section{The advantages of group quantization compared to standard per-layer or per-label quantization methods.}

\noindent \textbf{From the Perspective of Quantization Accuracy:} The primary challenge in quantization LLMs arises from the imbalanced numerical distribution (both between and within channels) and the prevalence of outliers in both weights and activations. Standard per-layer and per-token (or per-channel) quantization methods assume that the entire tensor or the neurons in each channel are identically distributed. This coarser quantization granularity is insufficient to address the issues of uneven distribution and outlier retention. Group quantization, however, further partitions the channels and quantizes the model weights at a finer granularity, effectively mitigating the problem of imbalanced numerical distribution and improving outlier handling, thereby reducing the accuracy loss typically associated with quantization.

\noindent \textbf{From the Perspective of Quantization Speed:} The finer quantization granularity of the per-group approach necessitates additional scaling factors during computation. However, since LLM tasks are memory-intensive rather than computation-bound, this increased granularity does not significantly impact memory access complexity compared to 2:4 sparsity. As a result, the inference speed is not adversely affected. For instance, the widely used reasoning engine, llama.cpp, employs group quantization for model inference.

\section{Results of GQSA on the Qwen model}

To verify the generalization ability of GQSA on different model families, we conduct experiments on Qwen models (base and instruct model). Table ~\ref{tab:ppl_qwen} shows similar results to the LLaMA model family, where GQSA consistently matches or outperforms the baseline methods even under stricter compression settings.

\section{Results of GQSA on the OPT model}

To verify the generalization ability of GQSA on different model families, we conduct experiments on OPT models (ranging from 1.3B to 13B parameters). Table ~\ref{tab:ppl_opt} shows similar results to the LLaMA and Qwen model family, where GQSA consistently matches or outperforms the baseline methods even under stricter compression settings.

\begin{table*}[ht]
\centering
\resizebox{\textwidth}{!}{%
\begin{tabular}{cccccccccc}
\toprule
 &  & \multicolumn{2}{c}{Qwen2.5-7B} & \multicolumn{2}{c}{Qwen2.5-14B} & \multicolumn{2}{c}{Qwen2.5-7B-Instruct} & \multicolumn{2}{c}{Qwen2.5-14B-Instruct} \\ \cmidrule(l){3-10} 
\multirow{-2}{*}{Setting} & \multirow{-2}{*}{Method} & WikiText2 & C4 & WikiText2 & C4 & WikiText2 & C4 & WikiText2 & C4 \\ \midrule
 & GPTQ & 29.22 & 89.12 & 23.08 & 55.43 & 46.03 & 289.92 & 48.42 & 38.37 \\
\multirow{-2}{*}{W2} & OmniQuant & 14.49 & 22.78 & 11.98 & 17.81 & 17.26 & 26.37 & 12.95 & 17.97 \\ \midrule
 & sparsegpt & \textbf{11.25} & 17.17 & 10.13 & \textbf{15.39} & 11.92 & 17.85 & 10.95 & 16.24 \\
\multirow{-2}{*}{2:4} & wanda & 14.78 & 22.84 & 11.74 & 18.24 & 15.80 & 23.83 & 12.06 & 18.73 \\ \midrule
\rowcolor[HTML]{EFEFEF} 
w4s20\% & \cellcolor[HTML]{EFEFEF} & 8.27 & 12.74 & 6.83 & 10.83 & 7.99 & 12.21 & 6.80 & 10.76 \\
\rowcolor[HTML]{EFEFEF} 
w4s30\% & \cellcolor[HTML]{EFEFEF} & 8.95 & 13.66 & 7.75 & 12.03 & 9.01 & 13.78 & 7.69 & 11.91 \\
\rowcolor[HTML]{EFEFEF} 
w4s40\% & \cellcolor[HTML]{EFEFEF} & 9.70 & 14.96 & 8.97 & 13.81 & 10.19 & 15.64 & 8.90 & 13.60 \\
\rowcolor[HTML]{EFEFEF} 
w4s50\% & \multirow{-4}{*}{\cellcolor[HTML]{EFEFEF}GQSA} & 11.71 & \textbf{17.02} & \textbf{9.87} & 15.93 & \textbf{11.74} & \textbf{17.07} & \textbf{10.81} & \textbf{15.97} \\ \bottomrule
\end{tabular}
}
\caption{Wikitext2 and C4 perplexity (↓) for Qwen2.5 models, with a context length of 2048.}
\label{tab:ppl_qwen}
\end{table*}

\begin{table*}[ht]
\centering
\resizebox{\textwidth}{!}{%
\begin{tabular}{cccccccccc}
\toprule
 &  & \multicolumn{2}{c}{OPT-1.3B} & \multicolumn{2}{c}{OPT-2.7B} & \multicolumn{2}{c}{OPT-6.7B} & \multicolumn{2}{c}{OPT-13B} \\ \cmidrule(l){3-10} 
\multirow{-2}{*}{Setting} & \multirow{-2}{*}{Method} & WikiText2 & C4 & WikiText2 & C4 & WikiText2 & C4 & WikiText2 & C4 \\ \midrule
 & GPTQ & 130.88 & 60.88 & 61.59 & 33.83 & 20.18 & 18.55 & 21.36 & 16.34 \\
 & QUIP & 41.64 & - & 28.98 & - & 18.57 & - & 16.02 & - \\
 & PB-LLM & 45.92 & - & 39.71 & - & 20.37 & - & 19.11 & - \\
 & OmniQuant & 23.95 & 27.33 & 18.13 & 21.11 & 14.43 & 16.67 & 12.94 & 14.92 \\
\multirow{-5}{*}{W2} & SliM-LLM & 24.57 & - & 17.98 & - & 14.22 & - & 12.16 & - \\ \midrule
 & SparseGPT & 24.54 & 26.55 & 17.82 & \textbf{19.45} & 14.23 & \textbf{16.56} & 12.94 & \textbf{14.88} \\
\multirow{-2}{*}{2:4} & Wanda & 28.27 & 28.54 & 21.17 & 22.84 & 15.90 & 18.99 & 15.55 & 16.18 \\ \midrule
W4S20\% & \cellcolor[HTML]{EFEFEF} & \cellcolor[HTML]{EFEFEF}14.49 & \cellcolor[HTML]{EFEFEF}16.60 & \cellcolor[HTML]{EFEFEF}12.03 & \cellcolor[HTML]{EFEFEF}14.54 & \cellcolor[HTML]{EFEFEF}10.21 & \cellcolor[HTML]{EFEFEF}12.71 & \cellcolor[HTML]{EFEFEF}9.93 & \cellcolor[HTML]{EFEFEF}12.16 \\
W4S30\% & \cellcolor[HTML]{EFEFEF} & \cellcolor[HTML]{EFEFEF}16.06 & \cellcolor[HTML]{EFEFEF}18.44 & \cellcolor[HTML]{EFEFEF}13.23 & \cellcolor[HTML]{EFEFEF}15.95 & \cellcolor[HTML]{EFEFEF}10.94 & \cellcolor[HTML]{EFEFEF}13.64 & \cellcolor[HTML]{EFEFEF}10.37 & \cellcolor[HTML]{EFEFEF}12.85 \\
W4S40\% & \cellcolor[HTML]{EFEFEF} & \cellcolor[HTML]{EFEFEF}18.82 & \cellcolor[HTML]{EFEFEF}21.54 & \cellcolor[HTML]{EFEFEF}15.39 & \cellcolor[HTML]{EFEFEF}18.25 & \cellcolor[HTML]{EFEFEF}12.15 & \cellcolor[HTML]{EFEFEF}15.12 & \cellcolor[HTML]{EFEFEF}11.29 & \cellcolor[HTML]{EFEFEF}13.97 \\
W4S50\% & \multirow{-4}{*}{\cellcolor[HTML]{EFEFEF}GQSA} & \cellcolor[HTML]{EFEFEF}\textbf{21.32} & \cellcolor[HTML]{EFEFEF}\textbf{24.90} & \cellcolor[HTML]{EFEFEF}\textbf{17.52} & \cellcolor[HTML]{EFEFEF}20.81 & \cellcolor[HTML]{EFEFEF}\textbf{13.44} & \cellcolor[HTML]{EFEFEF}16.94 & \cellcolor[HTML]{EFEFEF}\textbf{12.16} & \cellcolor[HTML]{EFEFEF}15.57 \\ \bottomrule
\end{tabular}
}
\caption{Wikitext2 and C4 perplexity (↓) for OPT models, with a context length of 2048.}
\label{tab:ppl_opt}
\end{table*}

\section{GQSA inference latency and memory consumption}
Due to space constraints, detailed inference latency and model memory consumption are provided in Appendix Table~\ref{tab:e2e_speed_appendix}. Overall, GQSA demonstrates exceptional performance across various settings.

\begin{table*}[ht]
\centering
\resizebox{\textwidth}{!}{%
\begin{tabular}{ccccccccc}
\toprule
\multicolumn{9}{c}{LLaMA-7B} \\ \midrule
 & \multicolumn{2}{c}{128} & \multicolumn{2}{c}{256} & \multicolumn{2}{c}{512} & \multicolumn{2}{c}{1024} \\ \cmidrule(l){2-9} 
\multirow{-2}{*}{sequence length} & Latency (ms) & Memory (GB) & Latency (ms) & Memory (GB) & Latency (ms) & Memory (GB) & Latency (ms) & Memory (GB) \\ \midrule
fp16 & 1490.5 & 13.47 & 3005.95 & 13.534 & 6090.97 & 13.662 & 12561.82 & 13.918 \\
w8a16 & 868.35 & 7.394 & 1755.62 & 7.458 & 3594.95 & 7.586 & 7559.22 & 7.842 \\
w8a16+sp0.3 & 688.89 & 6.296 & 1261.05 & 6.361 & 3005.02 & 6.489 & 5814.62 & 6.745 \\
w8a16+sp0.4 & 603.23 & 5.669 & 1103.08 & 5.733 & 2593.76 & 5.861 & 5039.33 & 6.117 \\
w8a16+sp0.5 & 512.71 & 5.042 & 996.59 & 5.106 & 2019.1 & 5.234 & 4329.32 & 5.492 \\ \midrule
w4a16 & 642.24 & 4.258 & 1312.91 & 4.322 & 2707.26 & 4.45 & 5786.8 & 4.706 \\
w4a16+g16+sp0.3 & 518.99 & 4.101 & 1041.18 & 4.165 & 2113.56 & 4.293 & 4437.48 & 4.549 \\
w4a16+g16+sp0.4 & 432.05 & 3.788 & 855.46 & 3.852 & 1828.48 & 3.977 & 3772.63 & 4.233 \\
w4a16+g16+sp0.5 & 377.98 & 3.474 & 699.26 & 3.528 & 1433.43 & 3.653 & 3110.54 & 3.909 \\ \midrule
\multicolumn{9}{c}{LLaMA-13B} \\ \midrule
 & \multicolumn{2}{c}{128} & \multicolumn{2}{c}{256} & \multicolumn{2}{c}{512} & \multicolumn{2}{c}{1024} \\ \cmidrule(l){2-9} 
\multirow{-2}{*}{sequence length} & Latency (ms) & Memory (GB) & Latency (ms) & Memory (GB) & Latency (ms) & Memory (GB) & Latency (ms) & Memory (GB) \\ \midrule
fp16 & 2726.66 & 25.61 & 5481.96 & 25.696 & 11071.81 & 25.92 & 22559.77 & 26.304 \\
w8a16 & 1439.66 & 13.524 & 2900.46 & 13.62 & 5922.28 & 13.844 & 12257.27 & 14.228 \\
w8a16+sp0.3 & 1164.239 & 11.396 & 2114.165 & 11.492 & 4976.471 & 11.716 & 9501.55 & 12.105 \\
w8a16+sp0.4 & 1024.199 & 10.182 & 1843.61 & 10.278 & 4272.72 & 10.502 & 8343.77 & 10.886 \\
w8a16+sp0.5 & 869.486 & 8.964 & 1715.976 & 9.061 & 3345.762 & 9.285 & 7044.25 & 9.669 \\ \midrule
w4a16 & 999.1 & 7.444 & 2020.99 & 7.54 & 4155.94 & 7.764 & 8750.98 & 8.148 \\
w4a16+g16+sp0.3 & 801.203 & 7.141 & 1475.175 & 7.237 & 3563.465 & 7.461 & 6972.12 & 7.845 \\
w4a16+g16+sp0.4 & 702.602 & 6.532 & 1303.865 & 6.628 & 3087.623 & 6.852 & 6081.292 & 7.236 \\
w4a16+g16+sp0.5 & 603.515 & 5.924 & 1104.366 & 6.02 & 2374.286 & 6.244 & 5099.638 & 6.628 \\ \midrule
\multicolumn{9}{c}{LLaMA-30B (TP=2)} \\ \midrule
 & \multicolumn{2}{c}{128} & \multicolumn{2}{c}{256} & \multicolumn{2}{c}{512} & \multicolumn{2}{c}{1024} \\ \cmidrule(l){2-9} 
\multirow{-2}{*}{sequence length} & Latency (ms) & Memory (GB) & Latency (ms) & Memory (GB) & Latency (ms) & Memory (GB) & Latency (ms) & Memory (GB) \\ \midrule
fp16 & 3759.08 & {\color[HTML]{1F2329} 65.534} & 7540.17 & {\color[HTML]{1F2329} 65.726} & 15241.36 & {\color[HTML]{1F2329} 66.11} & 31073.23 & {\color[HTML]{1F2329} 66.878} \\
w8a16 & 3032.64 & 32.418 & 6111.43 & 32.642 & 12371.66 & 33.026 & 25477.58 & 33.794 \\
w8a16+g16+sp0.3 & 2412.6 & 27.084 & 4861.575 & 27.308 & 10343.645 & 27.692 & 19826.459 & 28.46 \\
w8a16+g16+sp0.4 & 2132.65 & 24.036 & 3840.98 & 24.261 & 8925.685 & 24.645 & 17343.09 & 25.413 \\
w8a16+g16+sp0.5 & 1797.27 & 20.988 & 3472.16 & 21.212 & 6950 & 21.596 & 14591.638 & 22.364 \\ \midrule
w4a16 & 1938.2 & 17.178 & 3924.2 & 17.402 & 8011.57 & 17.786 & 16680.64 & 18.554 \\
w4a16+g16+sp0.3 & 1541.925 & 16.416 & 2515.512 & 16.641 & 6800.993 & 17.025 & 13290.836 & 17.793 \\
w4a16+g16+sp0.4 & 1341.315 & 14.892 & 2229.65 & 15.116 & 5890.861 & 15.501 & 11591.38 & 16.269 \\
w4a16+g16+sp0.5 & 1122.292 & 13.368 & 2180.11 & 13.592 & 4526.311 & 13.816 & 9720.279 & 14.584 \\ \bottomrule
\end{tabular}
}
\caption{Inference latency and memory usage of the FastTransformer implementation on NVIDIA A800-40GB GPU with a fixed input sequence length of 15, output sequence lengths of 128, 256, 512 and 1024.}
\label{tab:e2e_speed_appendix}
\end{table*}

\end{document}